\documentclass[10pt,twocolumn,twoside]{IEEEtran}

\usepackage{amsmath}
\usepackage{amssymb}
\usepackage{latexsym}
\usepackage{verbatim}
\usepackage{subfigure}
\usepackage[final]{graphicx}
\usepackage{psfrag}

\newtheorem{theorem}{Theorem}
\newtheorem{proposition}{Proposition}
\newtheorem{lemma}{Lemma}
\newtheorem{corollary}{Corollary}
\newtheorem{remark}{Remark}

\newtheorem{example}{Example}

{\begin{list}               
    {$\bullet$ \hfill}{
        \setlength{\leftmargin}{\parindent}
        \setlength{\parsep}{0.04\baselineskip}
        \setlength{\itemsep}{0.5\parsep}
        \setlength{\labelwidth}{\leftmargin}
        \setlength{\labelsep}{0em}}
    }
{\end{list}}

\providecommand{\eref}[1]{\eqref{eq:#1}}  
\providecommand{\cref}[1]{Chapter~\ref{chap:#1}}
\providecommand{\sref}[1]{Section~\ref{sec:#1}}
\providecommand{\fref}[1]{Figure~\ref{fig:#1}}

\providecommand{\R}{\ensuremath{\mathbb{R}}}

\providecommand{\abs}[1]{\lvert#1\rvert}
\providecommand{\norm}[1]{\lVert#1\rVert}
\providecommand{\inprod}[1]{\langle#1\rangle}
\providecommand{\set}[1]{\left\{#1\right\}}

\providecommand{\bydef}{\overset{\text{def}}{=}}

\renewcommand{\vec}[1]{\ensuremath{\boldsymbol{#1}}}
\providecommand{\mat}[1]{\ensuremath{\boldsymbol{#1}}}

\providecommand{\mQ}{\mat{Q}} \providecommand{\mR}{\mat{R}}
\providecommand{\mS}{\mat{S}}

 \providecommand{\mG}{\mat{G}}

\providecommand{\va}{\vec{a}}

 \providecommand{\vp}{\vec{p}}

\providecommand{\vx}{\vec{x}} \providecommand{\vy}{\vec{y}}

\providecommand{\vxi}{\vec{\xi}}

\usepackage{cite}

\usepackage{color}

\usepackage{booktabs,multirow}

\usepackage{enumitem}
\usepackage{commath}

\usepackage{algorithm, algorithmic}

\usepackage{hyperref}

\usepackage{esdiff}

\usepackage{pgf,tikz,psfrag}

\newcommand{\ie}{\emph{i.e.}, }
\newcommand{\eg}{\emph{e.g.}, }

\usepackage{dsfont}
\newcommand{\charfn}{\mathds{1}}

\newcommand{\argmin}{\operatornamewithlimits{argmin}}

\DeclareMathOperator{\prox}{prox}
\DeclareMathOperator{\normalize}{normalize}

\providecommand{\vxi}{\vec{\xi}}

\newcommand{\EE}{\mathbf{E}}
\newcommand{\PP}{\mathbf{P}}

\newcommand{\sgn}{\mathrm{sgn}}
\global\long\def\wt#1{\widetilde{#1}}
\global\long\def\f#1#2{\frac{#1}{#2}}

\global\long\def\nt{\left\lfloor nt\right\rfloor }
\global\long\def\ns{\left\lfloor ns\right\rfloor }
\global\long\def\a{\alpha}

\global\long\def\e{\epsilon}

\global\long\def\D{\Delta}

\def\O#1{\mathcal{O}(#1)}

\newcommand{\xki}{x_k^i}
\newcommand{\xnki}{x_{k+1}^i}
\newcommand{\xkj}{x_k^j}
\newcommand{\xko}{x_k^1}
\newcommand{\xii}{\xi^i}
\newcommand{\xij}{\xi^j}

\newcommand{\DMR}{D([0, T], \mathcal{M}(\R^2))}
\newcommand{\Dki}{\Delta_k^i}
\newcommand{\Dkj}{\Delta_k^j}
\newcommand{\Dko}{\Delta_k^1}
\newcommand{\Dkt}{\Delta_k^2}
\newcommand{\Gki}{\mathcal{G}_k^i}

\newcommand{\Lki}{\Lambda_k^i}

\newcommand{\gki}{g_k^i}
\newcommand{\gkj}{g_k^j}
\newcommand{\dki}{d_k^i}
\newcommand{\dkj}{d_k^j}
\newcommand{\zki}{z_k^i}
\newcommand{\zkj}{z_k^j}
\newcommand{\muk}{\mu_k}
\newcommand{\munk}{\mu_{k+1}}
\newcommand{\lfnt}{\lfloor nt \rfloor}

\begin{document}
%
\title{Scaling Limit: Exact and Tractable Analysis of Online Learning Algorithms with Applications to Regularized Regression and PCA}
\author{Chuang Wang, Jonathan Mattingly, and Yue M. Lu
\thanks{C. Wang is with the John A. Paulson School of Engineering and Applied Sciences, Harvard University, Cambridge, MA 02138, USA (e-mail: chuangwang@g.harvard.edu).}
\thanks{J. Mattingly is with the Department of Mathematics, Duke University, Durham, NC 27708, USA (e-mail: jonathan.mattingly@duke.edu).}
\thanks{Y. M. Lu is with the John A. Paulson School of Engineering and Applied Sciences, Harvard University, Cambridge, MA 02138, USA (e-mail: yuelu@seas.harvard.edu).  Part of this work was done during his semester-long visit to the Information Initiative at Duke (iiD) in Spring 2016. He thanks the members of this interdisciplinary program for their hospitality.}
\thanks{This work was supported in part by the ARO under contract W911NF-16-1-0265 and by the US National Science Foundation under grants CCF-1319140 and CCF-1718698.}
}

\markboth{}{Wang \MakeLowercase{et al.}: Exact and Tractable Analysis of Online Learning Algorithms}

\maketitle

\begin{abstract}
We present a framework for analyzing the exact dynamics of a class of online learning algorithms in the high-dimensional scaling limit. Our results are applied to two concrete examples: online regularized linear regression and principal component analysis. As the ambient dimension tends to infinity, and with proper time scaling, we show that the time-varying joint empirical measures of the target feature vector and its estimates provided by the algorithms will converge weakly to a deterministic measured-valued process that can be characterized as the unique solution of a nonlinear PDE. Numerical solutions of this PDE can be efficiently obtained. These solutions lead to precise predictions of the performance of the algorithms, as many practical performance metrics are linear functionals of the joint empirical measures. In addition to characterizing the dynamic performance of online learning algorithms, our asymptotic analysis also provides useful insights. In particular, in the high-dimensional limit, and due to exchangeability, the original coupled dynamics associated with the algorithms will be asymptotically ``decoupled'', with each coordinate independently solving a 1-D effective minimization problem via stochastic gradient descent. Exploiting this insight for nonconvex optimization problems may prove an interesting line of future research.
\end{abstract}


\section{Introduction}
\subsection{Motivations}

Many tasks in statistical learning and signal processing are naturally formulated as optimization problems. Examples include sparse signal recovery \cite{CandesRT:05, CandesT:06, Donoho:06}, principal component analysis (PCA) \cite{candes2011robust, dAspremont:2007zr}, low-rank matrix completion \cite{recht2010guaranteed, keshavan2010matrix}, photon-limited imaging \cite{duarte2008single, Harmany:2012uq}, and phase retrieval \cite{Candes:2013xy, Jaganathan:2013zl, Waldspurger:2015rz}. One distinctive feature of the optimization problems arising within such context is that we can often make additional statistical assumptions on their input and the underlying generative processes. These extra assumptions make it possible to study average performance over large random ensembles of such optimization problems, rather than focusing on individual realizations. Indeed, a long line of work \cite{Tanaka:2002fk, Guo:2007fj, donoho2009counting, donoho2009observed, Kabashima:2009qy, Rangan:2012lr, chandrasekaran2012convex, oymak2013squared, amelunxen2014living, javanmard2016phase} analyzed the properties of various optimization problems for learning and signal processing, predicting their exact asymptotic performances in the high-dimensional limit and revealing sharp phase transition phenomena. We shall refer to such studies as \emph{static} analysis, as they assume the underlying (usually convex) optimization problems have been solved and they characterize the properties of the optimizing solutions.

In this paper, we focus on the \emph{transient} performance---which we refer to as the \emph{dynamics}---of optimization algorithms. In particular, we present a tractable and asymptotically exact framework for analyzing the dynamics of a family of online algorithms for solving large-scale convex and nonconvex optimization problems that arise in learning and signal processing. In the modern data-rich regime, there are often more data than we have the computational resources to process. So, instead of iterating an algorithm on a limited dataset till convergence, we might be only able to run our algorithm for a finite number of iterations. Important questions to address now include the following: Given a \emph{fixed} budget on iteration numbers, what is the performance of our algorithm? More generally, what are the exact trade-offs between estimation accuracy, sample complexity, and computational complexity \cite{ChandrasekaranJ:13, BruerTCB:14, oymak2015sharp, giryes2016tradeoffs}?  When the underlying problem is non-stationary (as in adaptive learning and filtering), how well can the algorithms track the changing models? All these questions call for a clear understanding of the dynamics of optimization algorithms.


\vspace{-2ex}

\subsection{Examples of Online Learning Algorithms}
\label{sec:example_algorithms}

In this paper, we focus on two widely used algorithms, namely, regularized linear regression and PCA, and use them as prototypical examples to illustrate our analysis framework.

\begin{example}[Regularized linear regression]
Consider the problem of estimating a vector $\vxi \in \R^n$ from \emph{streaming} linear measurements of the form
\begin{equation}\label{eq:linear_model}
y_k = \tfrac{1}{\sqrt{n}}\va_k^T \vxi + w_k, \quad \text{for } k = 1, 2, \ldots.
\end{equation}
Here, we assume that the sensing vectors (or linear regressors) $\va_k \bydef [a_k^1, a_k^2, \ldots, a_k^n]^T$ consist of random elements $\set{a_k^i}$ that are i.i.d. over both $i$ and $k$. The noise terms $\set{w_k}$ in \eref{linear_model} are also i.i.d. random variables, independent of $\set{a_k^i}$. We further assume that  $\EE a_{k}^i = \EE w_k = 0$, $\EE (a_{k}^i)^2 = 1$, $\EE (w_k)^2 = \sigma^2$, and that all higher-order moments of $a_k^i$ and $w_k$ are finite. Beyond those moment conditions, we do not make further assumptions on the probability distributions of these random variables.

We analyze a simple algorithm for estimating $\vxi$ from the stream of observations $\set{y_k}$. Starting from some initial estimate $\vx_0$, the algorithm, upon receiving a new $y_k$, updates its estimate as follows:
\begin{equation}\label{eq:prox}
\vx_{k+1} = \eta\big[\vx_{k} + \tau (y_k - \tfrac{1}{\sqrt{n}} \va_k^T \vx_{k}) \tfrac{1}{\sqrt{n}}\va_k\big],
\end{equation}
for $k \ge 0$. Here, $\tau > 0$ is the learning rate, and $\eta(\cdot)$ is an element-wise (nonlinear) mapping taking the form
\begin{equation}\label{eq:eta}
\eta(x) = x - \tfrac{1}{n} \varphi(x),
\end{equation}
for some function $\varphi: \R \rightarrow \R$. Note that this is a streaming algorithm: it processes one sample at a time. Once a sample has been processed, it will be discarded and never used again. 

To see where the update steps \eref{prox} and the expression \eref{eta} come from, it will be helpful to first consider the following optimization formulation for regularized linear regression in the \emph{offline} setting:
\begin{equation}\label{eq:offline}
\widehat{\vx} = \underset{\vx}{\argmin} \ \frac{1}{2m} \sum_{k=1}^m (y_k - \tfrac{1}{\sqrt{n}} \va_k^T \vx)^2 + \frac{1}{n} \sum_{i=1}^n \Phi(x^i),
\end{equation}
where $m$ is the total number of data points used in the regression, $x^i$ denotes the $i$th element of $\vx$, and $\Phi(x)$ is a 1-D function providing a separable regularization term. For example, $\Phi(x) = \lambda \abs{x}$  corresponds to lasso-type penalizations; or we can choose $\Phi(x) = \lambda_1 x^2 + \lambda_2 \abs{x}$ for the elastic net \cite{Zou:2006}. When $\Phi(x)$ is convex, we can solve \eref{offline} using the proximal gradient method \cite{Parikh:2014}:
\begin{equation}\label{eq:proximal_regression}
\vx_{k+1}= \prox_{\Phi/n}\bigg[\vx_{k} + \frac{\tau}{m} \sum_{k \le m}   (y_k - \tfrac{1}{\sqrt{n}} \va_k^T \vx_{k}) \tfrac{1}{\sqrt{n}}\va_k\bigg],
\end{equation}
where $\prox_{\Phi/n}$ denotes the proximal operator of the function $\Phi(x) / n$. 

Replacing the full gradient in \eref{proximal_regression} by its instantaneous (and noisy) version $(y_k - \tfrac{1}{\sqrt{n}} \va_k^T \vx_{k}) \tfrac{1}{\sqrt{n}}\va_k$ and using the approximation $\prox_{\Phi/n}(x) \approx x - \tod{}{x}\!\Phi(x)/n$ (see, \emph{e.g.}, \cite[p. 138]{Parikh:2014} for a justification of this approximation which holds for large $n$), we reach our algorithm in \eref{prox} as well as the form given in \eref{eta}. Note that, when the regularizer $\Phi(x)$ is nonconvex, the proximal operator $\prox_{\Phi/n}$ is not well-defined. In this case, we can simply interpret \eref{prox} as a stochastic gradient descent method for solving \eref{offline}, with the function $\phi(x)$ in \eref{eta} chosen as $\tod{}{x}\!\Phi(x)$.

\end{example}

\begin{example}[Regularized PCA]
Suppose we observe a stream of i.i.d. $n$-dimensional sample vectors $\set{\vy_k}$ that are drawn from a distribution whose covariance matrix has a dominant eigenvector $\vxi \in \R^n$. More specifically, we assume the classical  spiked covariance model \cite{Johnstone:2001}, where the sample vectors are distributed according to
\begin{equation}\label{eq:spiked}
\vy_k = \sqrt{\frac{\omega}{n}} c_k \vxi + \va_k.
\end{equation}
Here, $\vxi \in \R^n$, with $\norm{\vxi} = \sqrt{n}$, is an unknown vector we seek to estimate, $\omega > 0$ is a parameter specifying the signal-to-noise ratio (SNR), and $\{c_{k}\}$ is a sequence of i.i.d. random variables with $\EE c_{k}=0$, $\EE c_{k}^{2}=1$ and finite higher-order moments. The assumption on the sequence of vectors $\{\va_{k}\}$ is the same as the one stated below \eref{linear_model}, and $\set{\va_k}$ are independent to $\set{c_k}$. It is easy to verify that $\vxi$ is indeed the leading eigenvector of the covariance matrix $\mat{\Sigma} = \EE \vy_k \vy_k^T$, and the associated leading eigenvalue is $1 + \omega$.

We study a simple streaming algorithm for estimating $\vxi$. As soon as a new sample $\vy_k$ has arrived, the algorithm updates its estimates of $\vxi$, denoted by $\vx_k$, using the following recursion
\begin{equation}\label{eq:oist}
\begin{aligned}
\widetilde{\vx}_{k} & = \vx_{k}+ \tfrac{\tau}{n} \, \vy_k \vy_k^T \vx_{k}\\
	\vx_{k+1}          & = \normalize\big[{\eta(\widetilde{\vx}_k)}\big].
\end{aligned}
\end{equation}
Here, $\tau > 0$ is the learning rate, $\eta(\cdot)$ is the same element-wise  mapping defined in \eref{eta}, and $\normalize(\vx)$ denotes the projection of $\vx$ onto the sphere of radius $\sqrt{n}$, \emph{i.e.}, $\normalize(\vx) \bydef \sqrt{n} \, \vx / \norm{\vx}$. Similar to our discussions in the previous example, the algorithm in \eref{oist} can be viewed as an online projected gradient method for solving the following optimization problem
\begin{equation}\label{eq:offline_pca}
\widehat{\vx} = \underset{\norm{\vx} = \sqrt{n}}{\argmin} \ \frac{-\vx^T \mat{\Sigma} \vx}{2n} + \frac{1}{n}\sum_{i=1}^n \Phi(x^i).
\end{equation}
In \eref{oist}, the full gradient $\mat{\Sigma} \vx / n$ is replaced by the noisy (but unbiased) approximation $\vy_k \vy_k^T \vx_k / n$, and $\eta(x)$ can be interpreted as the proximal operator of the regularizer (when the latter is convex) or simply as another gradient step taken with respect to $\Phi(x)$.

We note that, without the nonlinear mapping (\ie by setting $\eta(x) = x$), the recursions in \eref{oist} are exactly the classical Oja's method \cite{Oja:1985} for online PCA. The nonlinearity in $\eta(\cdot)$ can enforce additional structures on the estimates. For example, we can promote sparsity in the estimates by choosing $\varphi(x) = \beta \, \sgn(x)$ in \eref{eta}, which corresponds to adding an $L_1$-penalty $\Phi(x) = \beta \abs{x}$ in \eref{offline_pca}.

\end{example}

\subsection{Contributions and Paper Outline}
\label{sec:contributions}

In this paper, we provide an exact asymptotic analysis of the dynamics of the online regularized linear regression and PCA algorithms given in \eref{prox} and \eref{oist}. Specifically, let $\vx_k$ be the estimate of the target vector $\vxi$ given by the algorithm at the $k$th step. The central object of our study is the joint \emph{empirical measure} of $\vx_k$ and $\vxi$, defined as
\begin{equation}\label{eq:mu}
\mu_k^n(x, \xi) \bydef \frac{1}{n} \sum_{i= 1}^n \delta(x - x_k^i, \xi - \xi^i),
\end{equation}
where $x_k^i$ and $\xi^i$ denote the $i$th component of each vector, and the superscript in $\mu_k^n(x, \xi)$ makes the dependence on the ambient dimension $n$ explicit. Since $\vx_k$ is a random vector, $\mu_k^n(x, \xi)$ is a \emph{random} element in $\mathcal{M}(\R^2)$, the space of probability measures on $\R^2$. As the main result of this work, we show that, as $n \rightarrow \infty$ and with suitable time-rescaling\footnote{$k = \lfnt$, where $\lfloor \cdot \rfloor$ is the floor function. See \sref{toy_problem} for details.}, the sequence of random empirical measures $\set{\mu_k^n(x, \xi)}_n$ converges weakly to a deterministic measure-valued process $\mu_t(x, \xi)$. Moreover, this limiting measure $\mu_t(x, \xi)$ can be obtained as the unique solution of a nonlinear partial differential equation (PDE.)

The limiting measure provides detailed information about the dynamic performance of the algorithms, as many practical performance metric are just linear functionals of the joint empirical measures. For example, the mean squared error (MSE) is
\[
e_k \bydef  \frac{1}{n} \sum_i (\xi^i - \xki)^2 = \iint (\xi - x)^2 \mu_k^n(x, \xi) \dif x \dif \xi.
\]
More generally, any \emph{separable} loss function $L(\vx, \vxi) \bydef \tfrac{1}{n} \sum_i v(x^i, \xi^i)$ can be written as
\begin{equation}\label{eq:loss_function}
L(\vx_k, \vxi) = \iint v(x, \xi) \mu_k^n(x, \xi) \dif x \dif \xi.
\end{equation}

Since $\mu_k^n(x, \xi) \xrightarrow{n\to \infty} \mu_t(x, \xi)$, we can substitute the deterministic limiting measure $\mu_t$ for the (more challenging to handle) random empirical measure $\mu_k^n$ in \eref{loss_function} to obtain
\[
L(\vx_k, \vxi) \xrightarrow{n\to \infty} \iint v(x, \xi) \mu_t(x, \xi) \dif x \dif \xi,
\]
provided that the function $v(x, \xi)$ satisfies some mild technical conditions. (See Proposition~\ref{prop:par} for details.) We also note that our asymptotic characterization is tractable, as numerical solutions of the limiting PDE, which involves two spatial variables and one time variable, can be efficiently obtained by a contraction mapping procedure (see Remark~\ref{rem:numerical_PDE} in \sref{uniqueness}.)

A fully rigorous treatment of our asymptotic characterization requires some technical results related to the weak convergence of measure-valued stochastic processes. To improve readability, we organize the rest of the paper as follows. Readers who do not care about the full technical details can focus on \sref{results} and \sref{insights}. In \sref{results}, we state without proof the main results characterizing the asymptotic dynamics of regularized regression and PCA algorithms described in \sref{example_algorithms}. The validity and usefulness of these asymptotic predictions are demonstrated through numerical simulations. \sref{insights} presents the key ideas, including the important notion of \emph{finite exchangeability} \cite{Diaconis:1977, Diaconis:1980rr, aldous1985exchangeability}, behind our analysis. We also provide some insight obtained from our asymptotic analysis. In particular, in the high-dimensional limit and thanks to exchangeability, the original coupled dynamics associated with the algorithms will be asymptotically ``decoupled'', with each coordinate independently solving a 1-D effective minimization problem via stochastic gradient descent.  

Readers who are interested in applying our analysis framework to other algorithms should read \sref{formal}, where we present a \emph{formal} derivation leading to the limiting PDE for the case of regularized regression algorithm. Although they do not constitute a rigorous proof, the formal derivations can be especially convenient when one wants to quickly ``guess'' the limiting PDEs for the new algorithms. The rigorous theory of our analysis framework is fully developed in \sref{general}, where we establish a general weak-convergence ``meta-theorem'' characterizing the high-dimensional scaling and mean-field limit of a large family of stochastic algorithms that can be modeled as exchangeable Markov chain (see Theorem~\ref{thm:general}.) The online regularized regression and PCA algorithms considered in \sref{results} are just two special cases. In \sref{regression_proof}, we verify that all the assumptions of Theorem~\ref{thm:general} indeed hold for the case of online regression algorithm. Similar verifications can be done for regularized PCA, and we will report the details elsewhere. By considering the more general setting in the meta-theorem, we hope that the results of Theorem~\ref{thm:general} can be applied to establish the high-dimensional limits of other related algorithms beyond what we have covered in this paper.


%

\emph{Notation:} To study the high-dimensional limit of a stochastic iterative algorithm, we shall consider a sequence of problem instances, indexed by the ambient dimension $n$. Formally, we should use $\vx_k^{n}$ to denote an $n$-dimensional estimate vector provided by our algorithm at the $k$th iteration, and $x_{k}^{i, n}$ the $i$th coordinate of the vector. To lighten the notation, however, we shall omit the superscript $n$, whenever doing so causes no ambiguity. Let $\mu(x, \xi)$ be a probability measure on $\R^2$. For each function $f(x, \xi)$, we write
\[
\inprod{\mu, f} \bydef \iint f(x, \xi) \dif \mu(x, \xi).
\]
In particular, when $\mu(x, \xi) = \tfrac{1}{n} \sum_{i \le n} \delta(x - x^i, \xi - \xi^i)$ is a discrete measure concentrated on $n$ points $\set{(x^i, \xi^i)}_{i \le n}$, we have
\begin{equation}\label{eq:mu_f}
\inprod{\mu, f} = \frac{1}{n} \sum_i f(x^i, \xi^i).
\end{equation}
Using this notation, we can rewrite \eref{loss_function} as $L(\vx_k, \vxi) = \inprod{\mu_k^n, v}$.

\begin{figure*}[t]
        \centering
		\includegraphics[width=0.18\linewidth]{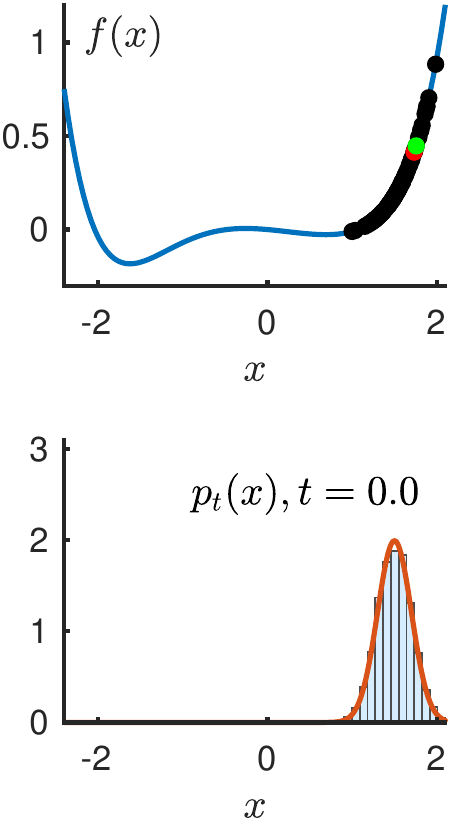}
		\hspace{1ex}
		\includegraphics[width=0.18\linewidth]{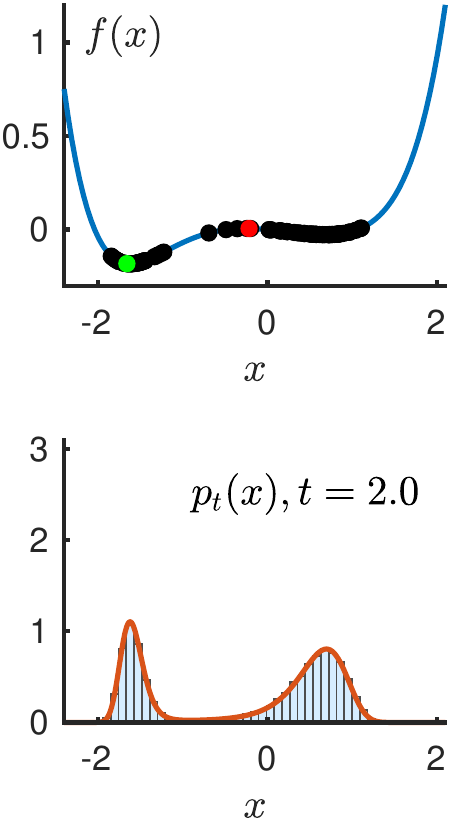}
		\hspace{1ex}
		\includegraphics[width=0.18\linewidth]{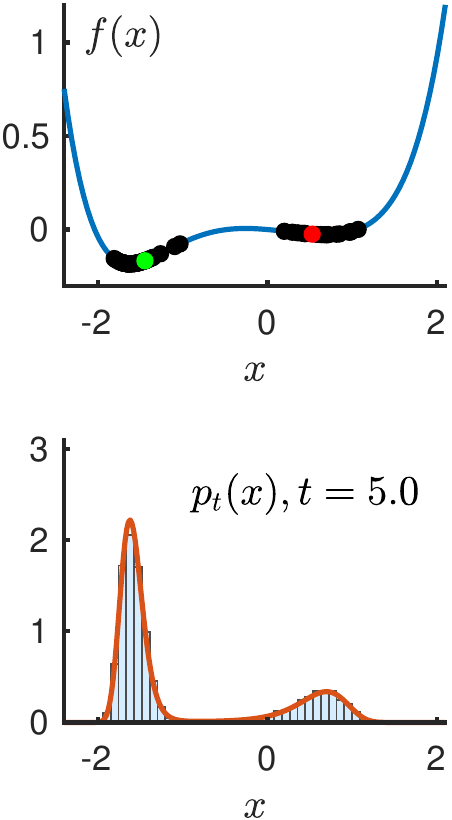}
		\hspace{1ex}
		\includegraphics[width=0.18\linewidth]{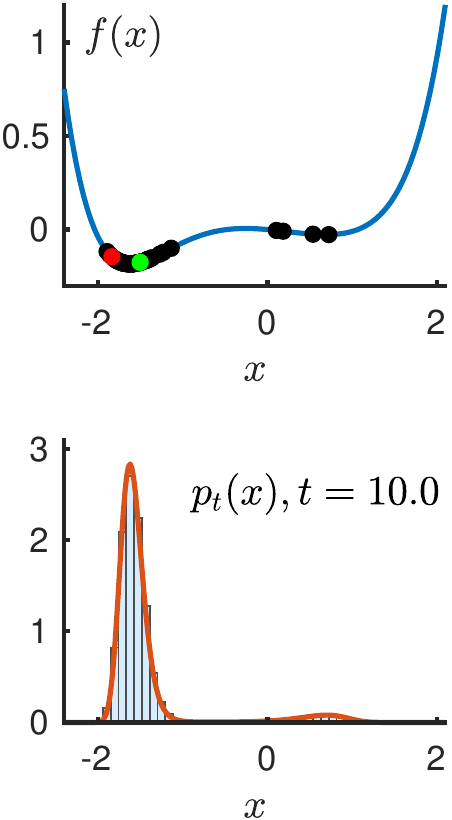}
		\hspace{1ex}
		\includegraphics[width=0.18\linewidth]{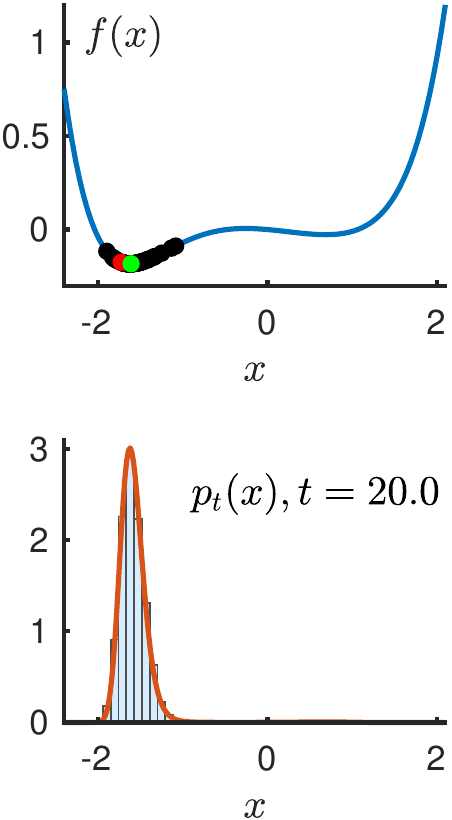}
 \caption{Time-varying probability densities of stochastic gradient descent for minimizing a 1-D nonconvex function. In the scaling limit, the densities $p_t(x)$ are the solution of a deterministic Fokker-Planck equation given in \eref{pde_1d}.}\label{fig:de}
 \end{figure*}

\section{Summary of Main Results for Regularized Regression and PCA}
\label{sec:results}

\subsection{Scaling Limits: A Toy Problem in 1-D}
\label{sec:toy_problem}

A common feature in our analysis is that a discrete-time stochastic process (\ie the algorithm we are studying) will converge, in some ``scaling limit'', to a continuous-time process that can be described by some PDE. Here, we use a simple \emph{one-dimensional} (1-D) example to illustrate some of the main ideas. Consider minimizing a 1-D function $f(x)$ shown in \fref{de} by using stochastic gradient descent:
\begin{equation}\label{eq:1d_proc}
x_k = x_{k-1} - \tfrac{\tau}{n} \, [f'(x_{k-1}) + \sqrt{n} \, v_k],
\end{equation}
where $\tau > 0$ is the learning rate, $v_k \overset{\text{iid}}{\sim} \mathcal{N}(0, \sigma^2)$ is a sequence of independent standard normal random variables, and $n > 0$ is a large constant introduced to scale the learning rate and the noise variance.\ (This particular choice of scaling is chosen here because it mimics the actual scaling that appears in the high-dimensional algorithms we study.) 

Suppose the initial iterand $x_0$ is drawn from a probability density $p_0(x)$. As the algorithm in \eref{1d_proc} is just a 1-D Markov chain, then in principle, the standard Chapman-Kolmogorov equation allows us to compute and track the exact \emph{evolution} of the probability densities $p_k(x)$ for the estimate $x_k$, at each step $k$. Computing the density evolution becomes easier in the (scaling) limit when $n \rightarrow \infty$. To see this, we first note that, when $n$ is large, the progress made by the algorithm between consecutive iterations is very small. In other words, we will not be able to see \emph{macroscopic} changes in $x_k$ unless we observe the process over a large number of steps. To \emph{accelerate} the time (by a factor of $n$), we embed $\set{x_k}$ in continuous-time by defining $x(t) \bydef x_{\lfloor nt \rfloor}$, where $\lfloor \cdot \rfloor$ is the floor function. Here, $t$ is the rescaled (accelerated) time: within $t \in [0, 1]$, the original algorithm proceeds $n$ steps. Standard results in stochastic processes \cite{Billingsley:1999, EthierK:85} show that the rescaled discrete-time process $\{x_{\lfnt}\}$ converges (weakly) to a continuous-time stochastic process $x(t)$ described by a drift-diffusion stochastic differential equation (SDE):
\[
\dif X_t = -\tau f'(X_t) \dif t + \tau \sigma \dif B_t,
\]
where $B_t$ denotes a standard Brownian motion. Let $p_t(x)$ denote the probability density of the solution of the SDE at time $t$. We can then apply standard results in the theory of SDE \cite{Oksendal:2003} to show that $p_t(x)$ is the unique solution of a \emph{deterministic} Fokker-Planck equation \cite{Risken:1996}
\begin{equation}\label{eq:pde_1d}
\tfrac{\partial}{\partial t} p_t(x) = \tfrac{\partial}{\partial x}\big[\tau f'(x) p_t(x)\big] + \tfrac{\tau^2 \sigma^2}{2} \tfrac{\partial^2}{\partial x^2} p_t(x).
\end{equation}

Solving numerically the above PDE then gives us the limiting probability densities $p_t(x)$ of the estimate $x(t)$ for all time $t$. Here, we demonstrate its usage through numerical simulations. In the second row of \fref{de}, we compare the limiting density $p_t(x)$ (shown as red solid lines) against empirical histograms (blue bars) formed by $1,000$ independent runs of the algorithm in \eref{1d_proc}, at 5 different time instants. The scaling parameter is chosen to be $n = 1,000$. Although the limit distributions are obtained in the asymptotic regime with $n \to \infty$, the results show that the theoretical predictions are accurate for a large but finite $n$. The figures also clearly show the ``migration'' of the estimates $x_k$ towards the global minimum as the algorithm progresses.

Knowing the probability densities $p_t(x)$ allows us to precisely characterize the \emph{dynamic} performance of the algorithm. Let $x^\ast$ be the global minimizer of $f(x)$, and $L(x, x^\ast)$ a general loss function. We can now compute, at the $k$th iteration, the expected loss $\EE\, L(x_k, x^\ast)$. We can also quantify the probabilities that the algorithm reaches the attraction basin of the global minimum, \ie $\PP(\norm{x_k - x^\ast} \le \delta)$ for some $\delta > 0$. Such questions are obviously important to both the designers and users of stochastic optimization algorithms.

For 1-D problems, the analysis described above is easy. But the challenge lies in performing the same analysis in high dimensions. In principle, we can still write out a PDE similar to \eref{pde_1d}, but the probability densities $p_t(\vx)$ will now be time-varying $n$-D functions where $n$ is the ambient dimension. As the dimension $n$ increases, it quickly becomes intractable to numerically solve such PDEs involving many variables.

\subsection{The Scaling Limit of Online Regularized Linear Regression}

In this section, we present the scaling limit that characterizes the asymptotic dynamics of the online regularized linear regression algorithm given in \eref{prox}. As mentioned earlier, the key object in our analysis is the joint empirical measure $\mu_k^n(x, \xi)$ of the estimate $\vx_k$ and the target $\vxi$, as defined in \eref{mu}. We note that $\mu_k^n$ is always a probability measure on $\R^2$, irrespective of the underlying dimension $n$. This is to be contrasted with the joint probability distribution $p(\vx_k, \vxi)$ of the two vectors, as the latter is function involving $2n$ variables.

To establish the scaling limit of $\mu_k^n$, we first embed the discrete-time sequence in continuous-time by defining
\begin{equation}\label{eq:mu_embedding}
\mu_t^n \bydef \mu_{\lfloor nt \rfloor}^n,
\end{equation}
just like what we did for the toy problem in \sref{toy_problem}. By construction, $\mu_t^n$ is a piecewise-constant c\`{a}dl\`{a}g process taking values in $\mathcal{M}(\R^2)$, the space of probability measures on $\R^2$. Since the empirical measures are random, $\mu_t^n$ is a random element in $D(\R^+, \mathcal{M}(\R^2))$, for which the notion of weak convergence is well-defined. (See, \emph{e.g.}, \cite{Kallenberg:2006} and our discussions in \sref{general}.)

In what follows, we state the asymptotic characterizations of the regularized regression algorithm. The assumptions on the sensing vectors $\set{\va_k}$ and the noise terms $\set{w_k}$ are the same moment conditions stated below \eref{linear_model}. In addition, we assume that the function $\varphi(x)$ in \eref{eta} is Lipschitz\footnote{The requirement that $\varphi(x)$ be a Lipschitz function is due to limitations of our current proof techniques. Numerical simulations show that our asymptotic predictions still hold when $\varphi(x)$ is piecewise Lipschitz, \eg $\varphi(x) = \beta\, \sgn(x)$.}. As a main result of our work, we show that, as $n \rightarrow \infty$, the sequence of time-varying empirical measures $\set{\mu_t^n(x, \xi)}_n$ converges weakly to a deterministic measure-valued process. It is stated formally as follows.

\begin{theorem}\label{thm:lasso}
Suppose that $\mu_0^n(x, \xi)$, the empirical measure for the initial vector $\vec{x}_0$ and the target vector $\vec{\xi}$, converges (weakly) to a deterministic measure $\mu_0 \in \mathcal{M}(\R^2)$ as $n \to \infty$. Moreover, $\sup_n \inprod{\mu_0^n, x^4 + \xi^4} < \infty$. Then, as $n \rightarrow \infty$, the measure-valued stochastic process $\mu_t^n(x, \xi)$ associated with the regularized regression algorithm converges weakly to a deterministic measure-valued process $\mu_t(x, \xi)$. Moreover, $\mu_t(x, \xi)$ is the unique solution to the following nonlinear PDE (given in the weak form): for any  positive, bounded and $C^3$ test function $f(x, \xi)$,
\begin{equation}\label{eq:pde_lasso_weak}
\begin{aligned}
\inprod{\mu_t, f} = \inprod{\mu_0, f} &+ \int_0^t \inprod{\mu_s, \left[\tau(\xi-x) - \varphi(x)\right]\tpd{}{x}\!f} \dif s \\
 &+ \frac{\tau^2}{2}\int_0^t (\sigma^2+e_s) \inprod{\mu_s, \tpd[2]{}{x}\!f} \dif s,
\end{aligned}
\end{equation}
where
\begin{equation}\label{eq:e_order}
e_t = \inprod{\mu_t, (x-\xi)^2}
\end{equation}
and $\varphi(x)$ is the function introduced in \eref{eta}.
\end{theorem}

\begin{remark}
The proof of this result is given in \sref{regression_proof}. The deterministic measure-valued process $\mu_t(x, \xi)$ characterizes the exact dynamics of the regularized regression algorithm in \eref{prox} in the high-dimensional limit. The nonlinear PDE \eref{pde_lasso_weak} specifies the time evolution of $\mu_t(x, \xi)$. Note that \eref{pde_lasso_weak} is presented in the weak form. If the strong, density valued solution\footnote{Here we slightly abuse notation by using $\mu_t(x, \xi)$ to denote both the probability measure and the associated probability density function.} exists, then it must satisfy the following strong form of the PDE:
\begin{equation}\label{eq:pde_lasso_strong}
\begin{aligned}
\tpd{}{t}\!\mu_t(x, \xi) &= \tpd{}{x}\! \bigg(-[ \tau(\xi-x) - \varphi(x)] \mu_t(x, \xi)\\
&\qquad\qquad+ \f{\tau^2}{2}[\sigma^2 + e(t)] \tpd{}{x}\!\mu_t(x, \xi) \bigg),
\end{aligned}
\end{equation}
where $e(t)$ is as defined in \eref{e_order}. The above PDE resembles the linear Fokker-Planck equation \eref{pde_1d} shown in \sref{toy_problem}. There is, however, one important distinction: the PDE \eref{pde_lasso_strong} involves a ``feedback'' term $e(t)$ that itself depends on the current solution $\mu_t$ as in \eref{e_order}. 
\end{remark}

In practice, one often quantifies the performance of the algorithm via various performance metrics, \eg MSE $e^n_t\bydef\inprod{\mu_t^n, (x-\xi)^2}$. The following proposition, whose proof can be found in \sref{meta}, shows that the asymptotic values of such performance metrics can be obtained from the limiting measure.

\begin{proposition} \label{prop:par}
Under the same assumptions of Theorem~\ref{thm:lasso}, we have
\begin{equation}\label{eq:weak_f}
\EE \abs{\inprod{\mu_t^n, f} - \inprod{\mu_t, f}} \xrightarrow[]{n\to \infty} 0,
\end{equation}
where $f(x, \xi)$ is any continuous function such that $\abs{f(x, \xi)} \le C(1+ x^2 + \xi^2)$ for some finite constant $C$.
\end{proposition}

As a special case of the above result, we have that $\EE \abs{e_t^n - e(t)} \xrightarrow{n\to\infty} 0$, where $e(t)$ is the function defined in \eref{e_order}.

\begin{figure}
\centering
\includegraphics[scale=0.8]{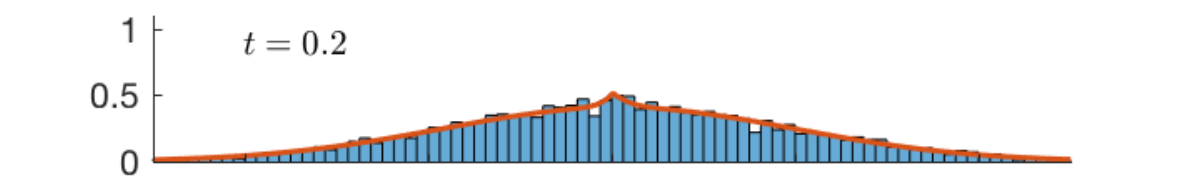}\\
\includegraphics[scale=0.8]{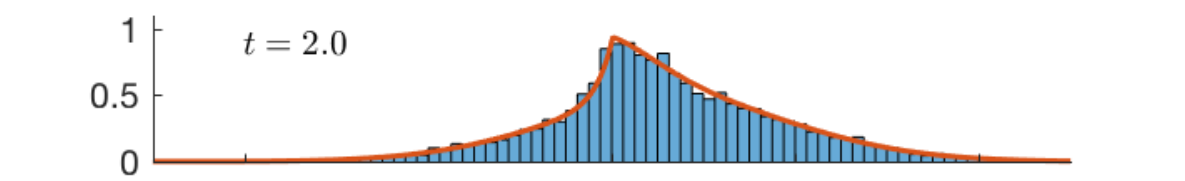}\\
\includegraphics[scale=0.8]{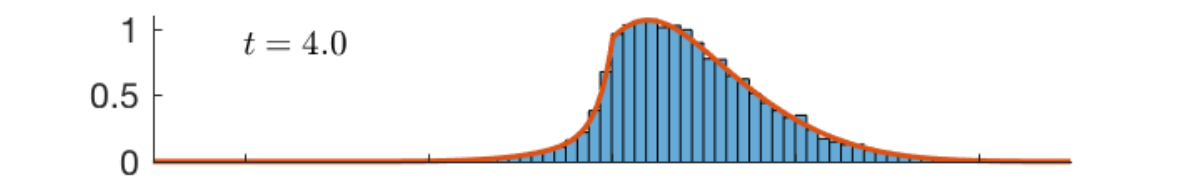}\\
\includegraphics[scale=0.8]{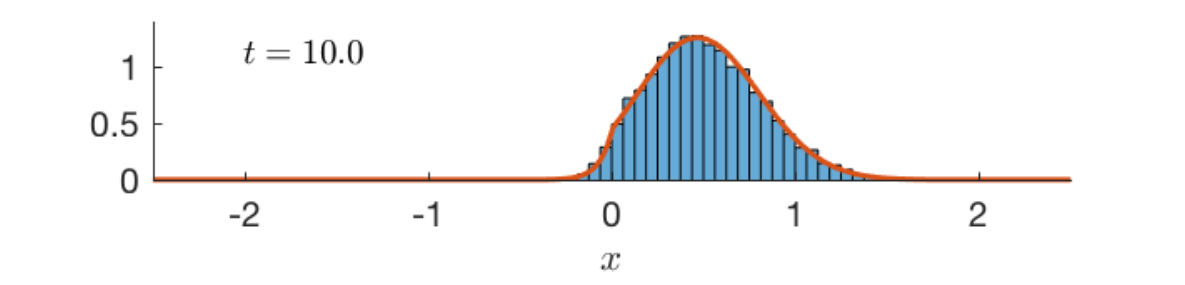}
\caption{\label{fig:pdf}
Asymptotic predictions v.s. simulations results. The red solid curves are predictions of the probability density $\pi_t(x \, \vert \, \xi=1)$ given by the PDE \eref{pde_lasso_strong}, and the blue bars show the empirical histograms of the estimates obtained by the online regularized regression algorithm at four different times. The signal dimension in this experiment is $n = 10^5$. 
}
\end{figure}

\begin{example}\label{example:lasso_sim}
We verify the accuracy of the theoretical predictions made in Theorem~\ref{thm:lasso} and Proposition~\ref{prop:par} via numerical simulations. In our experiment, we generate a vector $\vxi \in \R^n$ whose elements are either $1$ or $0$. The number of $1$s  is equal to to $\lfloor \rho n \rfloor$, where $\rho$ denotes the sparsity level. (The locations of the nonzero elements can be arbitrary.) Starting from a random initial estimate $\vx_0$ with i.i.d. entries drawn from the standard normal distribution, we use the online algorithm \eref{prox} to estimate $\vxi$. The nonlinear function $\varphi(x)$ in \eref{eta} is set to $\beta \sgn(x)$, which corresponds to using a regularizer $\Phi(x) = \beta \abs{x}$ in \eref{offline}. In our experiment, $\rho = 0.1$. The other parameters are $\tau = 0.2$, $\beta = 0.1$ and $\sigma = 1$. 

In \fref{pdf}, we compare the predicted limiting conditional densities $\pi_t(x \,\vert\, \xi = 1)$ against the empirical densities observed in our simulations, at four different times ($t = 0.2$, $t = 2.0$, $t = 4.0$ and $t = 10.0$.) The PDE in \eref{pde_lasso_strong} is solved numerically (see Remark~\ref{rem:numerical_PDE} in \sref{uniqueness}.) The comparison shows that that the limiting densities given by the theory provide accurate predictions for the actual simulation results. In \fref{mse_t}, we apply Proposition~\ref{prop:par} to predict the evolution of the MSE as a function of the time. For simulations, we average over $100$ independent instances of OIST, and plot the mean values and confidence intervals ($\pm 1$ standard deviations.) Again, we can see that the asymptotic results match with simulation data very well, even for moderate values of $n$.
\end{example}

\begin{figure}
\centering{}\includegraphics[scale=0.8]{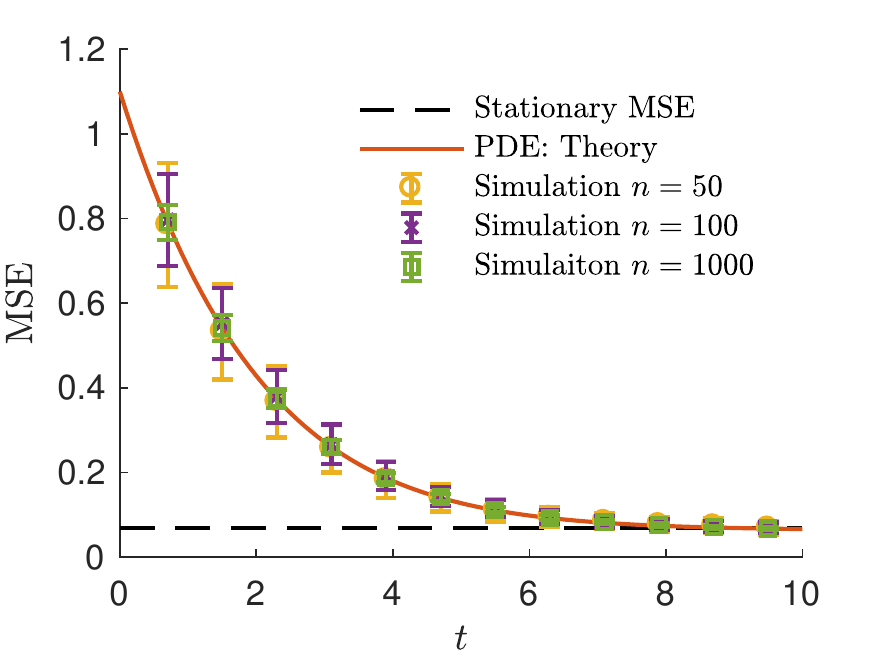}
\caption{\label{fig:mse_t}The mean square error (MSE) v.s. $t=k/n$: We run $100$ independent trials of the online learning algorithm for the regularized linear regression problem. The error bars show confidence intervals of one standard deviation. The result indicates that the empirical MSE curves converge to a deterministic one as $n$ increases. Moreover, this limit curve is accurately predicted by our asymptotic characterization.}
\end{figure}

\begin{figure}
\centering{}\includegraphics[scale=0.8]{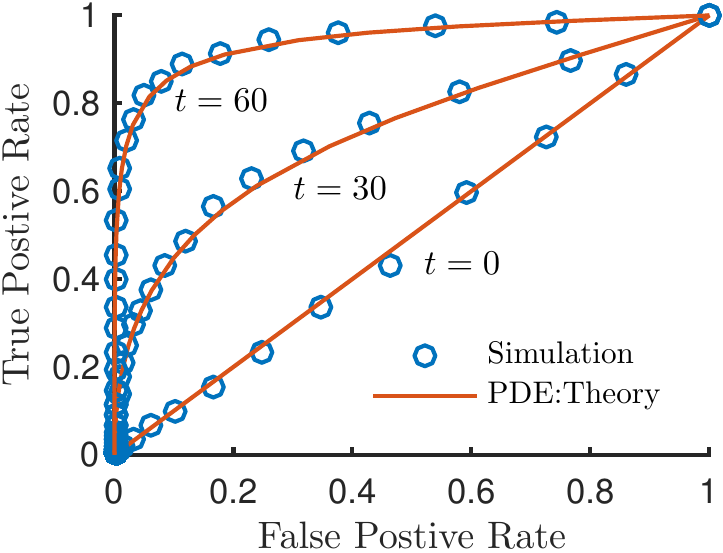}\caption{\label{fig:ROC}The trade-off between the true positive and false positive rates for sparse support estimation. The limiting measure as specified by the PDE \eref{pde_weak_pca} can accurately predict the exact trade-off at \emph{any} given time $t$.}
\end{figure}

\subsection{The Scaling Limit of Regularized PCA}

As one more example of our analysis framework, we present in what follows the high-dimensional scaling limit of the online regularized PCA algorithm described in \sref{example_algorithms}. Again, we study the joint empirical measure $\mu^n_k$ of the estimate $\vx_k$ and the target vector $\vxi$, and we use the time-rescaling \eref{mu_embedding} to define $\mu_t^n$. 

\begin{theorem} \label{thm:pca}
Under the same assumption on the initial empirical measure $\mu_0$ as stated in Theorem~\ref{thm:lasso}, the measure-valued process $\{\mu_t^n\}_{0\leq t \leq T}$ associated with the online regularized PCA algorithm \eref{oist} converges weakly to a deterministic measure-valued process $\{ \mu_t \}_{0\leq t \leq T}$. Moreover, $\mu_t$ is the unique solution of the following McKean-type PDE (given in the weak form): 
for any positive, bounded and $C^3$ test function $f(x,\xi)$,
\begin{equation}\label{eq:pde_weak_pca}
\begin{aligned}
\inprod{\mu_t, f} = \inprod{\mu_0, f} &+ \int_0^t \inprod{\mu_s, \mathcal{G}(x,\xi,Q_s,R_s) \tpd{}{x}\!f} \dif s \\
 &+ \frac{\tau^2}{2}\int_0^t (1+\omega Q_s^2) \inprod{\mu_s, \tpd[2]{}{x}\!f} \dif s,
\end{aligned}
\end{equation}
where
\begin{equation}
\begin{aligned}
\mathcal{G}(x,\xi,Q,R) &\bydef \tau \omega  Q \xi - \tau \beta \varphi(x)\\
 &\qquad- \tau x \big[ \omega Q - \beta R + \tfrac{\tau}{2} (1+\omega Q^2)\big]
\end{aligned}
\end{equation}
with
\begin{align}
Q_t &= \inprod{\mu_t, x\xi} \label{eq:pca_q}\\
R_t &=\inprod{\mu_t, x \varphi(x)},\label{eq:pca_r}
\end{align}
and $\varphi(x)$ is the function introduced in \eref{eta}.
\end{theorem}

\begin{remark}
The result of Proposition~\ref{prop:par} still holds for the online regularized PCA algorithm. (In fact, it is just a special case of a more general result shown in \sref{meta}.) Thus, if we define $Q_t^n = \inprod{\mu_t^n, x \xi} = \tfrac{1}{n}\sum x_{\lfloor nt \rfloor}^i \xi^i$, which measures the ``overlap'' or ``cosine similarity'' between $\vxi$ and the estimate $\vx_{\lfnt}$, we have $\EE \abs{ Q_t^n  - Q_t } \xrightarrow{n \to \infty} 0$, where $Q_t$ is the quantity defined in \eref{pca_q}. 
\end{remark}

\begin{example}[Support Recovery]
In the following example, we show that more involved questions, such as characterizing the misclassification rate in sparse support recovery, can also be answered by examining the limiting measure $\mu_t(x, \xi)$. Consider a sparse feature vector $\vxi$, consisting of $\lfloor \rho n \rfloor$ nonzero entries. For simplicity, we assume all the nonzero entries have the same value $1/\sqrt{\rho}$. This choice makes sure that $\norm{\vxi} = \sqrt{n}$. Starting from a random initial estimate $\vx_0$ with i.i.d. entries drawn from a normal distribution $\mathcal{N}(\tfrac{1}{\sqrt{2}}, \tfrac{1}{2})$, we use the regularized PCA algorithm \eref{oist}, with $\varphi(x) = \beta \,\sgn{x}$, to estimate $\vxi$. By applying a simple thresholding operation to $\vxi_k$, we estimate the support of $\vxi$ as
\[
\hat{s}_k^i = \charfn(\xki \ge c),
\]
where $c$ is a threshold parameter. This is to be compared against the true support pattern $s^i = \charfn(\xi^i \neq 0)$. The quality of the estimation $\hat{s}_k^i$ is measured by the true and false positive rates, the trade-off between which can be achieved by tuning the parameter $c$. Define $f_1(x, \xi) = \charfn(x \ge c, \xi = 1/\sqrt{\rho})$ and $f_2(x, \xi) = \charfn(x \ge c, \xi = 0)$. It is easy to see that the true and false positive rates can be computed\footnote{Technically, to apply the limit theorem as stated in \eref{weak_f}, the functions $f(x, \xi)$ must be continuous, but $f_1$ and $f_2$ defined here are only piecewise continuous. This restriction is imposed by the limitations of our current proof techniques. Numerical results in \fref{ROC} suggest that the expression \eref{weak_f} still holds for piecewise continuous functions.} as $\inprod{\mu_k^n, f_1}$ and $\inprod{\mu_k^n, f_2}$, respectively. In \fref{ROC}, we show that the limiting measure can be used to accurately predict the exact trade-off (i.e. the ROC curve) in support recovery, at \emph{any} given time $t$. Being able to analytically predict such performance is valuable in practice, as users will know the exact number of iterations (or the number of samples) they need in order to achieve a given accuracy.
\end{example}

\subsection{Connections to Prior Work}

 Analyzing the convergence rate of stochastic gradient descent has already been the subject of a vast literature \cite{nemirovski1982, Benveniste:90, KushnerY:03, nemirovski2009robust, rakhlin2011making, shalev2011pegasos, roux2012stochastic, iouditski2014primal}. Unlike existing work which studies the problem in \emph{finite dimensions}, we analyze the performance of stochastic algorithms in the \emph{high-dimensional limit}. Moreover, we explicitly explore the extra assumptions on the generative models for the observations, which allow us to analyze the exact limiting dynamics of the algorithms.
 
The basic idea underlying our analysis can trace its root to the early work of McKean \cite{mckean1967propagation, mckean1966class}, who studied the statistical mechanics of Markovian-type mean-field interactive particles. In the literature, this line of research is often known under the colorful name \emph{propagation of chaos}, whose rigorous mathematical foundation was established in the 1980's (see, \eg \cite{Meleard:1987, Sznitman:1991}.) Since then, related ideas and tools have been successfully applied to problems outside of physics, including numerical PDEs \cite{Bossy:1997}, game theory \cite{aumann1964markets}, particle filters \cite{ristic2004beyond}, and Markov chain Monte Carlo algorithms \cite{roberts1997weak}. As we show in our preliminary results, a large family of stochastic iterative algorithms can be viewed as mean-field interactive particle systems, but this perspective does not seem to have been taken before our recent work \cite{WangL:16, WangL:17}.


Our goal of tracking the time-varying probability densities of the algorithms bears resemblance to density evolutions in the analysis of LDPC decoding \cite{Richardson:2008} and general message passing on graphical models \cite{Mezard:2009}. In particular, closely related to our work is the \emph{state evolution} analysis of the approximate message passing (AMP) algorithm \cite{Donoho:2009, Bayati:2011, Rangan:2010, vila2013expectation, rangan2016vector}. The standard AMP is a block (\ie parallel update) algorithm. Due to the introduction of the \emph{Onsager reaction term} \cite{Donoho:2009, Bayati:2011} into the recursion, the probability densities associated with AMP are asymptotically Gaussian, and thus the states of the algorithm can be followed by tracking a collection of \emph{scalar parameters}. In this work, we study a broad family of stochastic iterative algorithms that may not necessarily have the Onsager term built-in. Moreover, the algorithms we consider have low-complexity updating rules, where each step might only use one measurement. Due to these features, the probability densities associated with our algorithms are not in parametric forms (see, \eg \fref{pdf}), and thus our proposed analysis requires more detailed \emph{density} evolution using PDEs.

\section{Main Ideas and Insights}
\label{sec:insights}

\subsection{Exchangeable Distributions}
\label{sec:exchangeability}

Finite \emph{exchangeability} \cite{Diaconis:1977, Diaconis:1980rr, aldous1985exchangeability} is a key ingredient that allows us to perform exact analysis of high-dimensional stochastic processes in a tractable way. It is also heavily used in our technical derivations in \sref{general} and \sref{regression_proof}. A joint distribution $p(x_1, x_2, \ldots, x_n)$ of $n$ random variables is exchangeable, if
\[
p(x_1, x_2, \ldots, x_n) = p(x_{\pi(1)}, x_{\pi(2)}, \ldots, x_{\pi(n)})
\]
for any permutation $\pi$ and for all $x_1, x_2, \ldots, x_n$. A simple example of an exchangeable distribution is $p(x_1, x_2, \ldots, x_n) = \prod_{1 \le i \le n} p(x_i)$, \ie when the coordinates $\set{x_i}$ are i.i.d. random variables. The family of exchangeable distributions, however, is much larger than i.i.d. distributions. (See \cite{Diaconis:1977} for an interesting example showing the geometry of finite exchangeable distributions.)


The concept of exchangeability naturally extends to Markov chains defined on $\mathcal{S}^{\otimes n}$, where $\mathcal{S}$ is a Polish space\footnote{For the purpose of our subsequent discussions, it is sufficient to consider the special case where $\mathcal{S} = \R^2$.}. Let $\Pi_n$ be the set of permutations of $\set{1, 2, \ldots, n}$. For a given permutation $\pi \in \Pi_n$ and a given $\vx = (x_1, x_2, \ldots, x_n)^T \in \mathcal{S}^{\otimes n}$, define
\[
\pi \circ \vx = (x_{\pi(1)}, x_{\pi(2)}, \ldots, x_{\pi(n)})^T.
\]
Similarly, for any subset $\mathcal{B} \subset \mathcal{S}^{\otimes n}$, we have $\pi \circ \mathcal{B} = \set{\pi \circ \vx: \vx \in \mathcal{B}}$. Consider a homogeneous Markov chain with kernel
\[
K(\vx, \mathcal{B}) \bydef \PP(\vx_{k+1} \in \mathcal{B} \mid \vx_{k} = \vx).
\]
We say that a Markov chain is exchangeable, if
\begin{equation}\label{eq:kernel}
K(\vx, \mathcal{B}) = K(\pi \circ \vx, \pi \circ \mathcal{B})
\end{equation}
 for arbitrary $\vx \in \mathcal{S}^{\otimes n}$, Borel set $\mathcal{B}$, and permutation $\pi$. 

It is important to note that the online regularized regression and PCA algorithms defined in \eref{prox} and \eref{oist} can both be seen as exchangeable Markov processes on $(\R^2)^{\otimes n}$. To see this, we focus on the regularized regression algorithm. At each step $k$, the state of this Markov chain is $[(x_k^1, \xi_k^1), (x_k^2, \xi_k^2), \ldots, (x_k^n, \xi_k^n)]$, or equivalently, a pair of vectors $(\vx_k, \vxi_k)$. The states of the Markov chain evolve according to the following dynamics:
\begin{equation}\label{eq:lasso_exchangeable}
\begin{cases}
\vx_{k+1} = \eta\big[\vx_{k} + \tau ( \tfrac{1}{\sqrt{n}} \va_k^T (\vxi_k - \vx_{k}) + w_k) \tfrac{1}{\sqrt{n}}\va_k\big]\\
\vxi_{k+1} = \vxi_{k},
\end{cases}
\end{equation}
where we have substituted the observation model \eref{linear_model} into \eref{prox}. The second update equation $\vxi_{k+1} = \vxi_{k}$ is trivial, due to the fact that the target vector $\vxi$ stays fixed. However, in order to establish exchangeability, it is convenient to include $\vxi_k$ as part of the state of the Markov chain. 

Next, we verify that \eref{lasso_exchangeable} is indeed an exchangeable Markov chain. Note that any permutation $\pi \in \Pi_n$ can be represented by a matrix $\pi \in \R^{n \times n}$, where the $i$th row of $\pi$ is $e_{\pi(i)}^T$. With this notation, we can write $\pi \circ \vx = \pi \vx$, where the right-hand side is a standard matrix-vector product. To verify the exchangeability condition \eref{kernel}, we first note that
\[
\begin{aligned}
&\PP((\vx_{k+1}, \vxi_{k+1}) \in \pi \circ \mathcal{B} \, \mid \, (\pi \circ \vx_k, \pi \circ \vxi_{k-1}))\\
&\qquad\qquad= \PP((\pi^T \vx_{k+1}, \pi^T \vxi_{k+1}) \in B\, \mid \, (\pi  \vx_k, \pi  \vxi_{k-1})).
\end{aligned}
\] 
Since
\[
\begin{aligned}
\pi^T \vx_{k+1} &= \eta\big[\pi^T \pi \vx_k + \tau ( \tfrac{1}{\sqrt{n}} \va_k^T (\pi \vxi_k - \pi \vx_{k}) + w_k) \tfrac{1}{\sqrt{n}}\pi^T \va_k\big]\\
	&= \eta\big[\vx_k + \tau ( \tfrac{1}{\sqrt{n}} (\pi^T\va_k)^T (\vxi_k - \vx_{k}) + w_k) \tfrac{1}{\sqrt{n}}\pi^T \va_k\big]
\end{aligned}
\]
and due to the exchangeability of the random vector $\va_k$, we conclude that \eref{kernel} holds and that \eref{lasso_exchangeable} is indeed an exchangeable Markov chain. Using similar arguments, we can also show that the regularized PCA algorithm in \eref{oist} is an exchangeable process.

Exchangeability has several important consequences that are key to our analysis framework. First, given an exchangeable Markov chain in $(\R^2)^{\otimes n}$, the corresponding empirical measure $\mu_k(x, \xi)$ as defined in \eref{mu} forms a Markov chain in the space of probability measures $\mathcal{M}(\R^2)$. This simple fact is known in the literature. To make our discussions self-contained, we provide a simple proof in Appendix~\ref{appendix:useful_lemma}. Thanks to this property, we just need to study the evolution and the scaling limit of this measure-valued process, irrespective of the underlying dimension $n$. The essence of our asymptotic analysis boils down to the following: as $n$ increases, this measure-valued process ``slows down'', and the stochastic process converges, after time-rescaling, to a deterministic measure-valued process whose evolutions are exactly described by the limiting PDEs given in Theorem~\ref{thm:lasso} and Theorem~\ref{thm:pca}.

Second, it is easy to verify the following property of exchangeable Markov chains: if the initial state $(\vx_0, \vxi_0)$ is drawn from an exchangeable distribution, then for any $k > 0$, the probability distribution of $(\vx_k, \vxi_k)$ remains exchangeable. In the context of our online learning algorithms, this property means that, at any step $k$, different coordinates of the estimate $\vx_k$ are statistically symmetric. This then allows us to simplify a lot of our technical derivations given in \sref{general}.

\begin{remark}
The requirement that the initial state $(\vx_0, \vxi_0)$ must be exchangeable seems to be an overly restrictive condition, as it makes a strong statistical assumption on the target vector $\vxi$. Fortunately, this restriction can be removed by using a simple ``trick''. Let $(\vx_0, \vxi_0)$ be any \emph{deterministic} initial state for the algorithm. We first apply a \emph{random} permutation, drawn uniformly from $\Pi_n$, to $(\vx_0, \vxi_0)$. The resulting permuted state $(\pi \circ \vx_0, \pi \circ \vxi_0)$ becomes exchangeable, and it is then used as the new initial state for the Markov chain. Although the subsequent outputs $(\vx_k, \vxi)$ resulting from this new initial state will be different from the actual outputs from the original algorithm, one can see that the sequence of empirical measures $\mu_k^n$ of the two versions of the algorithms have exactly the same probability distributions. Thus, since we only seek to characterize the asymptotic limit of the empirical measures, we can assume without loss of generality that the initial state has been randomly permuted, and hence exchangeable.
\end{remark}

Another very important consequence of exchangeability is the following. If $(\vx, \vxi)$ is drawn from an exchangeable distribution, and if the empirical measure $\mu^n(x, \xi)$ converges weakly to a deterministic measure $\mu(x, \xi)$ as $n \to \infty$, then for any \emph{finite} integer $m$, the joint probability distribution on the first $m$ coordinates of $\vx$ and $\vxi$ converges to a factorized distribution. More specifically,
\begin{equation}\label{eq:asymptotic_fact}
p(x^1, \xi^1, x^2, \xi^2, \ldots, x^m, \xi^m) \xrightarrow[]{n\rightarrow \infty} \prod_{i\le m} \mu(x^i, \xi^i).
\end{equation}
See, \eg \cite{Sznitman:1991} for a proof. Moreover, due to exchangeability, the same property holds for \emph{any} subset of $m$ coordinates. Roughly speaking, \eref{asymptotic_fact} implies that different coordinates of the vectors $\vx$ and $\vxi$ will be \emph{asymptotically independent}.


\begin{figure*}[h]
\begin{subfigure}[ ]
{\includegraphics[scale=0.47]{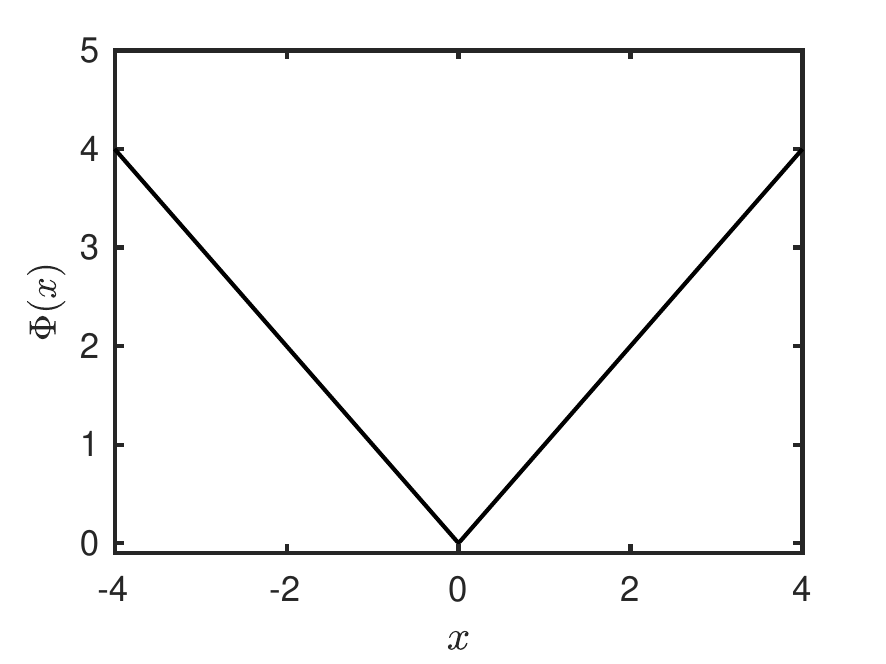}}
\end{subfigure}
\begin{subfigure}[ ]
{\includegraphics[scale=0.47]{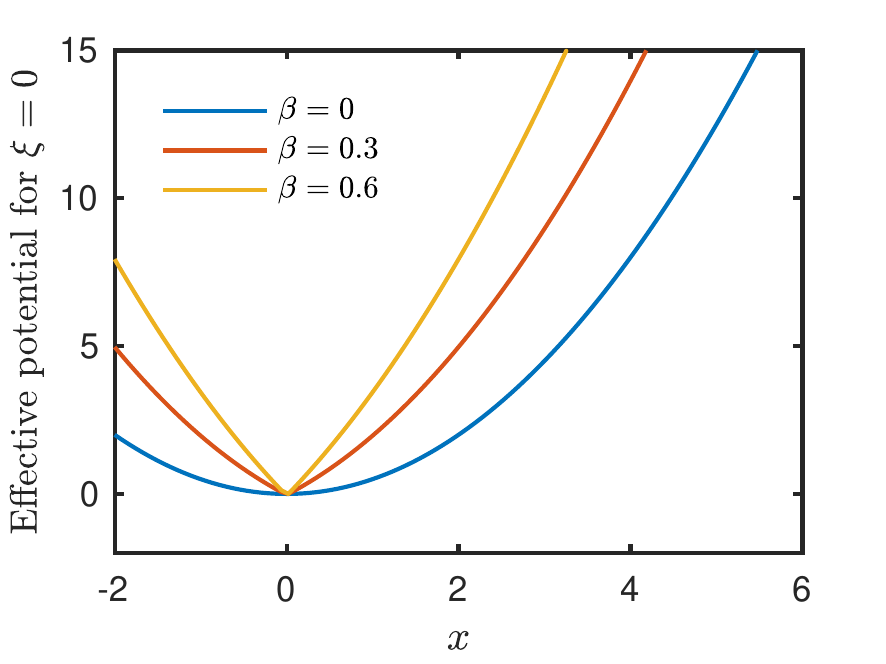}}
\end{subfigure}
\begin{subfigure}[ ]
{\includegraphics[scale=0.47]{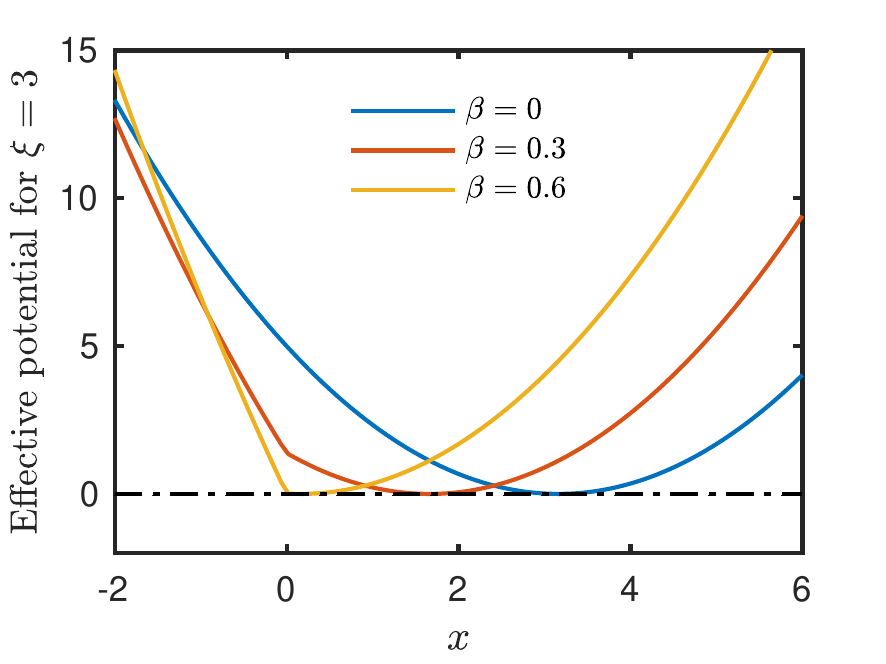}}
\end{subfigure}
\begin{subfigure}[ ]
{\includegraphics[scale=0.47]{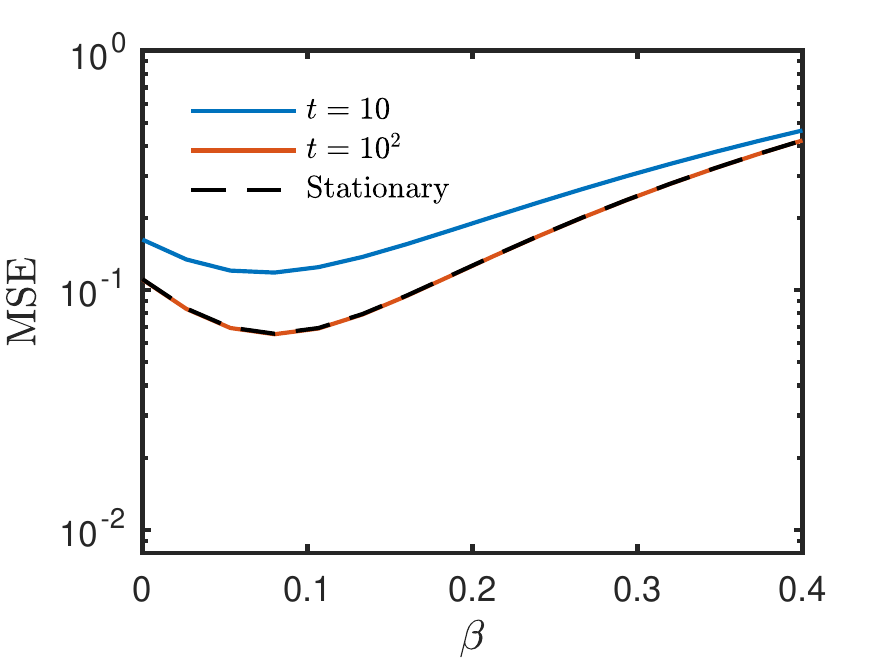}}
\end{subfigure}
\\
\begin{subfigure}[$\;$]
{\includegraphics[scale=0.47]{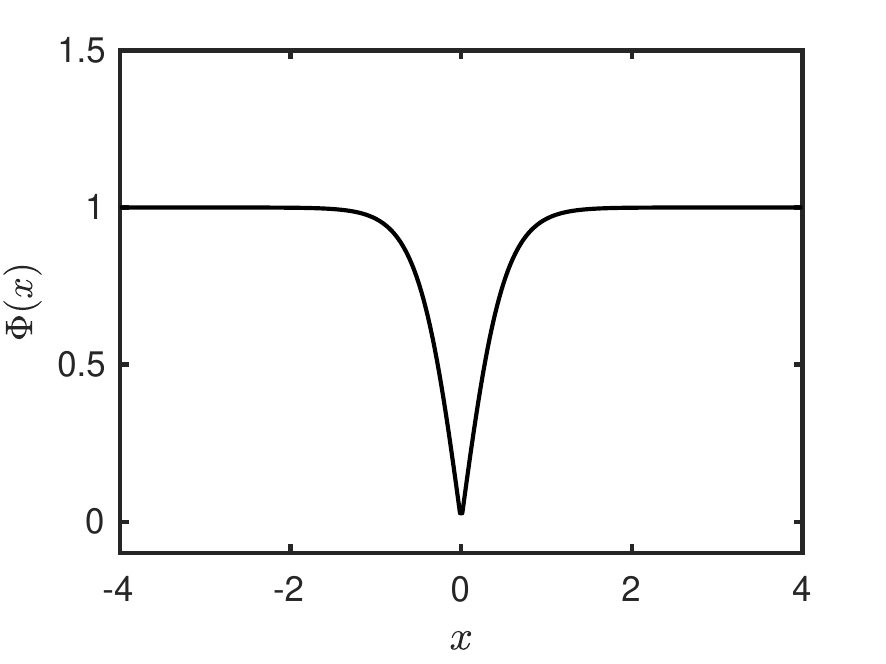}}
\end{subfigure}
\begin{subfigure}[$\;$]
{\includegraphics[scale=0.47]{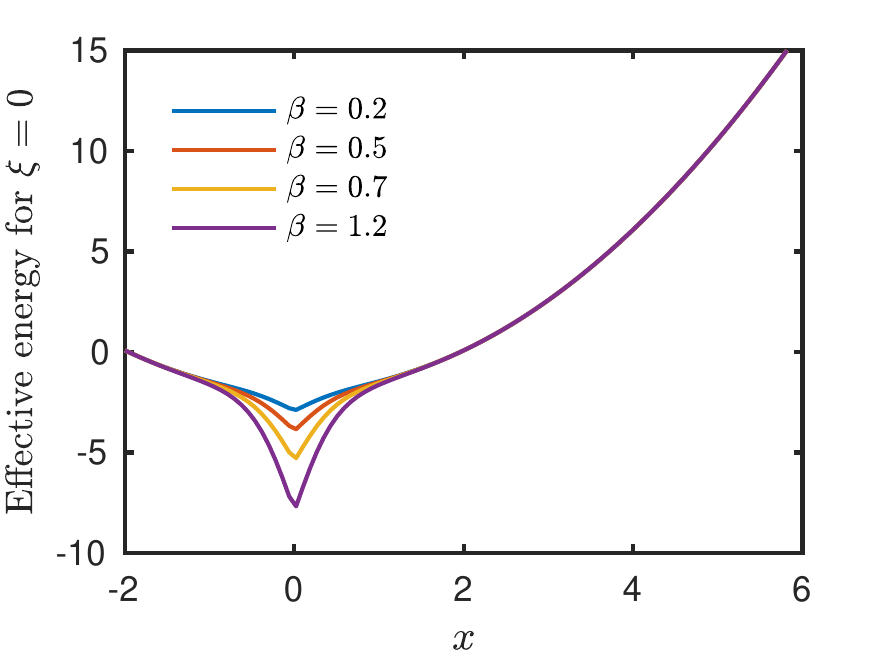}}
\end{subfigure}
\begin{subfigure}[$\;$]
{\includegraphics[scale=0.47]{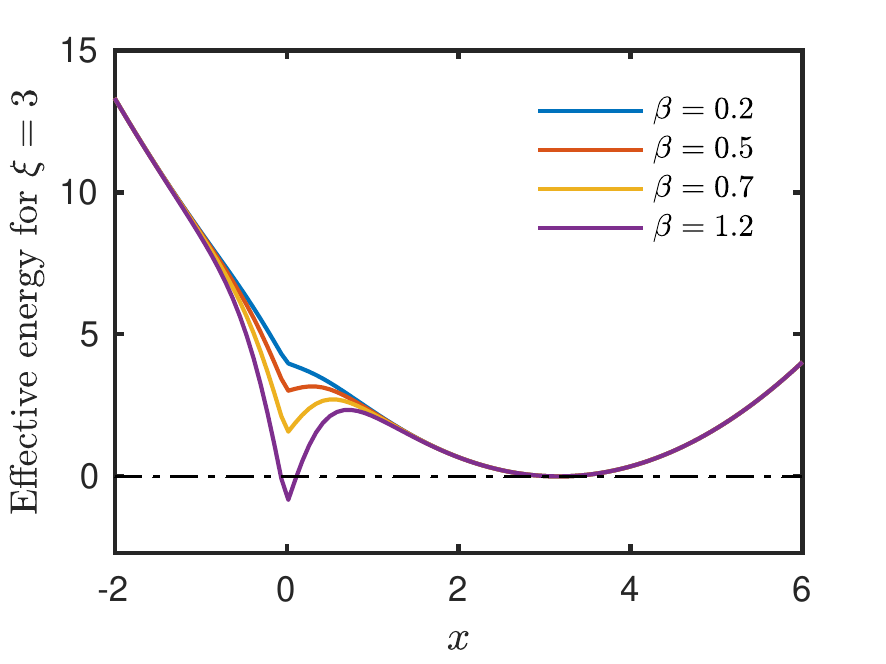}}
\end{subfigure}
\begin{subfigure}[$\;$]
{\includegraphics[scale=0.47]{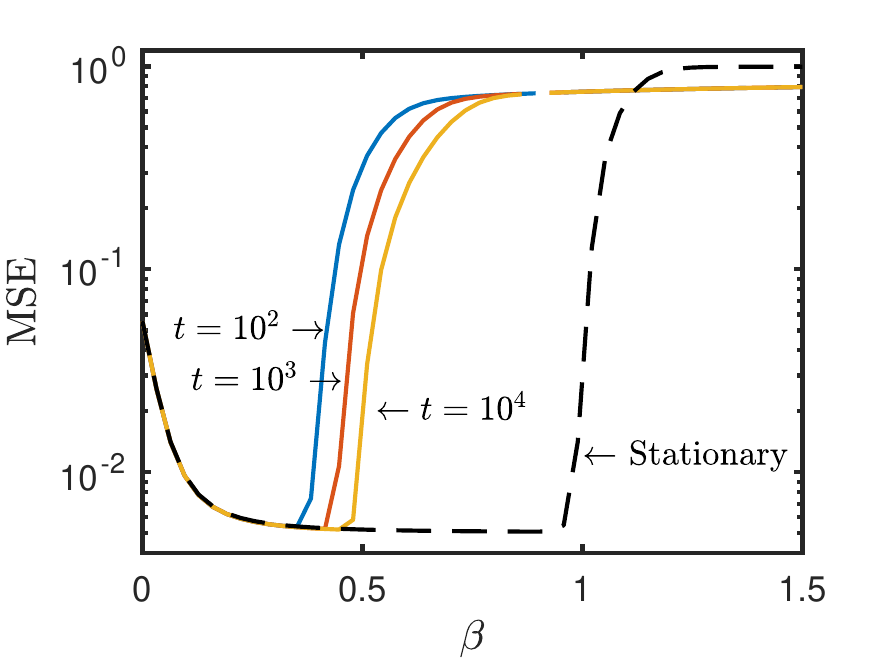}}
\end{subfigure}
\caption{\label{fig:1d}
Dynamics of the regularized regression algorithm using a convex regularizer $\Phi(x)=|x|$ (the first row) and a nonconvex regularizer $\Phi(x)=\tanh(\alpha\left|x\right|)$ (the second row). The first column shows the two regularizers. The second and third columns show the effective 1-D potential \eqref{eq:opt-1d}
for $\xi=0$ and $\xi = 3$, respectively. The fourth column shows the MSE v.s. regularization strength $\beta$ at different iteration times $t$. The learning rate $\tau=0.2$ is fixed.  
The figures show a nonconvex regularizer may have a better performance. However, inappropriate 
algorithmic parameters can make the dynamics trapped in a metastable state for a very long time
when another local minimum emerges in the 1-D potential function.}
\end{figure*}

\subsection{Insights: Asymptotically Uncoupled Dynamics}
\label{sec:uncoupled}

In what follows, we show how the property \eref{asymptotic_fact} leads to some useful insights regarding the online regression and PCA algorithms. First, these algorithms correspond to exchangeable Markov chains. Moreover, our asymptotic characterizations given in Theorem~\ref{thm:lasso} and Theorem~\ref{thm:pca} state that the empirical measures associated with the algorithms indeed converge to some deterministic limits. It then follows from \eref{asymptotic_fact} that, in the large $n$ limit, the coupled processes associated with the online learning algorithms become effectively \emph{uncoupled}. The dynamics of each coordinate $(x_k^i, \xi_k^i)$, as we will show next, resembles a stochastic gradient descent solving a $1$-D effective optimization problem.

We explain this interpretation using the example of online regression, whose asymptotic characterization is given by the limiting PDE \eref{pde_lasso_strong}. Note that this PDE involves a feedback term $e(t)$ as defined in \eref{e_order}. However, if $e(t)$ were given in an \emph{oracle} way, then \eref{pde_lasso_strong} would just be a standard Fokker-Planck equation, describing the evolution of the time-marginal distribution of a drift-diffusion process
\begin{equation}
\dif X_{t}=[ \tau(\xi- X_t) - \varphi(X_t)] \dif t+\tau\sqrt{\sigma^2+e(t)} \dif B_{t}.
\label{eq:sde}
\end{equation}
There is an interesting interpretation of \eref{sde}: Suppose $\varphi(x) = \beta \tod{}x\!\Phi(x)$, where $\Phi(x)$ is the regularization function used in \eref{offline}, and define a 1-D effective potential function 
\begin{equation}
E_{\xi}(x)=\tfrac{\tau}{2}(x-\xi)^{2}+\beta\Phi(x).\label{eq:opt-1d}
\end{equation}
Since the drift term $[ \tau(\xi- X_t) - \varphi(X_t)]$ of \eref{sde} is exactly the negative gradient of $E_{\xi}(x)$, the SDE \eref{sde} can be viewed as a \emph{continuous-time} stochastic gradient descent for minimizing this effective potential function.

\begin{example}
In this example, we use the above insight to study how a nonconvex regularizer $\Phi(x)$ can affect the dynamical performance of the online regression algorithm. Similar to our setting in Example~\ref{example:lasso_sim}, we consider a sparse target vector $\vxi$ whose entries are either $0$ or $3$, and the sparsity level is denoted by $\rho$. We consider two regularizers: a convex one $\Phi(x)=\left|x\right|$ and a nonconvex one 
\[
\Phi(x)=\tanh(\alpha \abs{x})
\]
with $\alpha = 2$ [see \fref{1d}(e)]. Fixing the learning rate as $\tau=0.2$, we study the dynamics of the algorithm using different regularization strength $\beta$.  Figure~\ref{fig:1d}(b) and Figure~\ref{fig:1d}(f) plot the effective potential function
$E_{\xi}(x)$ for $\xi=0$, for the convex and nonconvex regularizers, respectively.  We can see that, as $\beta$ increases, 
the effective potential grows a deeper and deeper valley at $x=0$, thus forcing the estimates $x$ to concentrate around 0.

Figure~\ref{fig:1d}(c) and Figure~\ref{fig:1d}(g) plot the corresponding effective potentials 
$E_{\xi}(x)$ for $\xi=3$.  For the convex regularizer shown in Figures \ref{fig:1d}(c), the effective potential is always a convex function.
Thus, as we can see in Figure \ref{fig:1d}(d), the algorithm can quickly converge to its stationary state for all $\beta$ values. A drawback of using the convex regularizer is that it will shift the minimum of the effective potential towards
$0$, and the resulting bias can lead to a larger stationary error. In contrast, the nonconvex regularizer applies a more ``gentle'' penalization for larger estimate errors. It thus has a smaller bias, resulting a lower stationary error. However, if the regularization strength $\beta$ is above a certain threshold, another local minimum of the potential function will emerge at $x=0$, and the algorithm dynamics will be trapped in this local
minimum for a very long time.  Such phenomenon is reflected in Figure \ref{fig:1d}(h), where the dynamics is still far from reaching its stationary state, even for $t=10^{4}$. In the statistical physics literature, this is known as a metastable state in double-well potential systems. The presence of metastability suggests that, when a nonconvex regularizer is used, various algorithmic parameters such as $\beta$ and $\tau$ must be chosen more carefully than in the convex case. Moreover, the optimal choices of these parameters critically depend on the target run time of the algorithms [as seen in \fref{1d}(h).]
\end{example}

\section{Formal Derivations of the Limiting PDEs}
\label{sec:formal}

In this section, we focus on the example of online regularized regression \eref{prox}, and present a convenient \emph{formal} approach that allows us to quickly derive the limiting PDE \eref{pde_lasso_weak}. Rigorously establishing this asymptotic characterization will be done in \sref{general} and \sref{regression_proof}.

\subsection{Moment Calculations}
\label{sec:moment}

Let $x_k^i$ denote the $i$th element of $\vx_k$. Using \eref{linear_model}, we rewrite the algorithm in \eref{prox} as
\begin{equation}\label{eq:lasso}
\xnki = \eta\big[\xki + \tau (w_k - \tfrac{1}{\sqrt{n}} \va_k^T (\vx_{k} - \vxi)) \tfrac{1}{\sqrt{n}}a_k^i\big].
\end{equation}
By introducing
\begin{equation}\label{eq:gki}
g_k^i = \tau (w_k - \tfrac{1}{\sqrt{n}} \va_k^T (\vx_{k} - \vxi)) \tfrac{1}{\sqrt{n}}a_k^i,
\end{equation}
and
\begin{equation}\label{eq:Dki_full}
\Delta_k^i = \gki - \frac{\varphi(\xki + \gki)}{n}
\end{equation}
we can write \eref{lasso} as
\begin{equation}\label{eq:Dki}
\xnki = \xki + \Delta_k^i.
\end{equation}
For each $k \ge 0$, we denote by $\mathcal{F}_k^n$ the sigma field generated by $(x_0^{1}, x_0^{2}, \ldots, x_0^{n})$, $(\xi_0^{1}, \xi_0^{2}, \ldots, \xi_0^{n})$ and $(a_{\ell}^{1}, a_{\ell}^{2}, \ldots, a_{\ell}^{n}, w_\ell)_{0 \le \ell \le k-1}$. Throughout the rest of the paper, we use $\EE_k(\cdot)$ as a shorthand notation for the conditional expectation $\EE(\cdot \, \vert \, \mathcal{F}_k^n)$.

Straightforward derivations give us
\begin{align}
\EE_{k} \, g_k^i &= \frac{-\tau}{n} (\xki - \xi^i)\label{eq:g_m1}\\
\EE_{k} \, (g_k^i)^2 &= \frac{\tau^2}{n} (\sigma^2 + e_k)  + \frac{\tau^2(\EE (a^i)^4 - 1)}{n^2} (\xki - \xi^i)^2 \label{eq:g_m2}\\
\EE_{k} \, g_k^i g_k^j &=  \frac{2\tau^2}{n^2} (\xki - \xi^i)(\xkj - \xi^j), \qquad i \neq j.\label{eq:g_ij}
\end{align}
In \eref{g_m2}, $e_k$ is the MSE at the $k$th step, defined as
\begin{equation}\label{eq:MSE}
e_k \bydef \frac{1}{n}\sum_{1 \le i \le n} (\xki - \xi^i)^2.
\end{equation}
Using these calculations and the Lipschitz property of $\varphi(x)$, we can estimate the first and second (conditional) moments of the difference term $\Dki$ as follows:
\begin{equation}\label{eq:m1_leading}
\EE_k \Dki =  -\frac{1}{n}[\tau(x-\xi) + \varphi(x)] + \mathcal{O}(n^{-3/2})
\end{equation}
and
\begin{equation}\label{eq:m2_leading}
\EE_k (\Dki)^2 =  \frac{\tau^2(\sigma^2 + e_k)}{n} + \mathcal{O}(n^{-3/2}).
\end{equation}
(See Lemma~\ref{lemma:m1m2} and Proposition~\ref{prop:x4e2} in \sref{regression_proof} for more precise statements.) 

The expressions \eref{m1_leading} and \eref{m2_leading} provide the leading-order terms of the drift and diffusion components of the discrete-time stochastic process $\set{\xki}_k$. The scaling $(1/n)$ that appears in both expressions also indicate that $(1/n)$ is indeed the \emph{characteristic} time of the process, and hence it makes sense to rescale the time as $k = \lfnt$, which is what we do in our asymptotic analysis.  

\subsection{Decompositions and Embeddings}

Our goal is to derive the asymptotic limit of the time-varying empirical measures $\set{\mu_t^n(\xi, x)}_n$. The convergence of empirical measures can be studied through their actions on test functions. Let $f(x, \xi)$ be a nonnegative, bounded and $C^3$ test function. Recall the definition of $\inprod{\mu, f}$ as given in \eref{mu_f}. From the general update equation \eref{Dki} and using Taylor-series expansion,
\begin{equation}\label{eq:expansion}
\begin{aligned}
&\inprod{\munk, f} - \inprod{\muk, f} = \frac{1}{n} \sum_i \tpd{}{x}f(\xki, \xii) \Dki \\
&\qquad\qquad\quad+ \frac{1}{2n} \sum_i \tpd[2]{}{x}f(\xki, \xii) (\Dki)^2 + r_k,
\end{aligned}
\end{equation}
where $r_k = \frac{\tau^3}{6n}\sum_i \tpd[3]{}{x}f(c_k^i, \xi_k^i) (\Dki)^3$, with $c_k^i \in [\xki, \xki + \Dki]$, collects the higher-order remainder terms.

Introducing two sequences
\[
v_k \bydef \EE_k(\inprod{\munk, f} - \inprod{\muk, f} - r_k) 
\]
and
\begin{equation}\label{eq:mk}
m_k \bydef \inprod{\munk, f} - \inprod{\muk, f} - r_k - v_k,
\end{equation}
we have
\[
\inprod{\munk, f} - \inprod{\muk, f} = v_k + m_k + r_k
\]
and thus
\[
\begin{aligned}
\inprod{\mu_k,f} - \inprod{\mu_{0},f} &= \sum_{0 \le \ell < k} v_\ell + \sum_{0 \le \ell < k} m_\ell + \sum_{0 \le \ell < k} r_\ell \\
&= V_k + M_k + R_k,
\end{aligned}
\]
where
\[
V_k \bydef \sum_{0 \le \ell < k} v_\ell, \ M_k \bydef \sum_{0 \le \ell < k} m_\ell, \ R_k \bydef \sum_{0 \le \ell < k} r_\ell, 
\]
for $k \ge 1$. We also set $V_0 = M_0 = R_0 = 0$. By construction, the sequence $\set{M_k}$ is a martingale null at 0


From \eref{m1_leading} and \eref{m2_leading}, we have
\begin{align}
v_k &=  \frac{1}{n^2} \sum_i \tpd{}{x}f(\xki, \xii) \EE_k \Dki \nonumber\\
&\qquad\qquad+ \frac{1}{2n^2} \sum_i \tpd[2]{}{x}f(\xki, \xii)\EE_k(\Dki)^2  + \O{n^{-3/2}}\nonumber\\
&= \frac{1}{n} \inprod{\muk,  \left[\tau(\xi-x) - \varphi(x)\right]\tpd{}{x}\!f}\nonumber\\
&\qquad\qquad+ \frac{\tau^2(\sigma^2 + e_k)}{2n}\inprod{\muk, \tpd[2]{}{x}\!f} + \O{n^{-3/2}}.\label{eq:v2}
\end{align}

Embedding the discrete sequence $V_k$ in continuous-time and speeding up time by a factor of $n$, we define 
\[
V(t) \bydef V_{\lfloor nt \rfloor}, 
\]
which is a piecewise-constant c\`{a}dl\`{a}g function in $D([0, T], R)$. Similarly, we can define $M(t), R(t), \mu_t$ as the continuous-time rescaled versions of their discrete-time counterparts. Since $V(t)$ is piecewise-constant over intervals of length $1/n$, the expression in \eref{v2} can be written as
\[
v_k = \int_{\frac{k}{n}}^{\frac{k+1}{n}} L(\mu_s) \dif s + \O{n^{-1/2}},
\]
where
\begin{equation}\label{eq:L_gen}
\begin{aligned}
L(\mu_s) &\bydef  \inprod{\mu_s, \left[\tau(\xi-x) - \varphi(x)\right]\tpd{}{x}\!f}\\
&\qquad\qquad\qquad+ \frac{\tau^2(\sigma^2 + e_s)}{2}\inprod{\mu_s, \tpd[2]{}{x}f}. 
\end{aligned}
\end{equation}
It follows that
\begin{align}
&\inprod{\mu_t, f} - \inprod{\mu_0, f} = V(t) + M(t) + R(t)\nonumber\\
&\qquad=\int_0^t L(\mu_s) \dif s + M(t) + R(t) + \O{n^{-1/2}}.\label{eq:term_by_term}
\end{align}

We note that $L(\mu_s)$ contains exactly the last two terms on the right-hand side of the limiting PDE \eref{pde_lasso_weak}. \emph{Formally}, if the martingale term $M(t)$ and the higher-order term $R(t)$ converge to $0$ as $n \to \infty$, we can then establish the scaling limit. A rigorous proof, however, requires a few more ingredients, following a standard recipe in the literature \cite{Meleard:1987, Sznitman:1991}. The details will be given in the next sections.


\section{The Scaling Limits of General Exchangeable Markov Chains}
\label{sec:general}

In this section, we consider a general family of exchangeable Markov chains of which the online algorithms for the two example problems in \sref{example_algorithms} are  special cases. We present and prove a meta-theorem, establishing the scaling limits for the time-varying empirical measures associated with these Markov chains.

\emph{Notation}: Throughout the rest of the paper, we use $C$ to denote a generic and finite constant that does not depend on $n$ or $k$. To streamline the derivations, the exact values of $C$ can change from line to line. Similarly, for any $T > 0$, the notation $C(T)$ represents a generic and finite constant whose value can depend on $T$ but not on $n$ or $k$. We also use $x \vee y$ and $x \wedge y$ to denote $\max\set{x, y}$ and $\min\set{x, y}$, respectively. Finally, for any $b \in (0, \infty)$ and $x \in \R \cup \set{\pm \infty}$, we use
\begin{equation}\label{eq:proj_B}
x \sqcap b = (\,\abs{x} \wedge b) \, \sgn(x)
\end{equation}
to denote the projection of $x$ onto the interval $[-b, b]$. When $\vx$ is a vector, $\vx \sqcap b$ represents the element-wise application of \eref{proj_B} to $\vx$.

\subsection{The Meta-Theorem}
\label{sec:meta}

Consider a Markov chain $(\vx_k, \vxi_k)$ defined on $(\R^2)^{\otimes n}$. The states of the Markov chain can be represented by two $n$-dimensional vectors $\vec{\xi_k}$ and $\vx_{k}$. 
Moreover, we assume that $ \vec{\xi}_{k} \equiv \vec{\xi}$ for all $k$, and thus we will omit the subscript $k$ in $\vxi_k$ to emphasize that $\vxi_k$ is invariant. 



We first state the basic assumptions under which our results are proved.
\begin{enumerate}[label={(A.\arabic*)}]
\item \label{a:exchangeable} The Markov chain $\set{(\vx_{k}, \vec{\xi})}_{k \ge 0}$ is exchangeable. 

\item The initial empirical measure $\mu_0^n$ converges weakly to a deterministic measure $\mu_0^\ast$.

\item There is some finite constant $C$ such that
\[
\sup_n \, \inprod{\mu_0^n, x^4 + \xi^4} \le C.
\]

\item \label{a:drift} Let $ \Dki=\xnki-\xki$. There exists a deterministic function $\mathcal{G}: \R \times \R \times \R^r \mapsto \R$, for some $r \ge 0$, such that, for each $T > 0$,
\[
\max_{k \le nT} \EE \abs{ \EE_{k} \Dki- \tfrac{1}{n} \Gki }  \le \frac{C(T)}{n^{1+\gamma}},
\]
where $\gamma > 0$ is some positive constant. In the above expression,
\begin{equation}\label{eq:Gki}
\Gki = \mathcal{G}(\xki, \xii, \vec{Q}_k^n)
\end{equation}
and $\vec{Q}_k^n = [Q_k^n(1), Q_k^n(2), \ldots, Q_k^n(r)]$ is an $r$-dimensional vector. The $\ell$th element of $\vec{Q}_k^n$ is defined as
\[
Q_{k}^{n}(\ell)= \inprod{\mu_k^n, p_\ell(x, \xi)},
\]
where $p_\ell(x, \xi)$ is some deterministic function.

\item \label{a:diffusion} There exists a deterministic function $ \Lambda: \R^{r} \mapsto \R$ such that, for each $T > 0$,
\[
\max_{k \le nT} \EE \abs{ \EE_{k} (\Dki)^{2}- \tfrac{1}{n} \Lambda_k } \le \frac{C(T)}{n^{1+\gamma}},
\]
where $\gamma > 0$ is some positive constant, and  
\begin{equation}\label{eq:Lki}
\Lambda_k= \Lambda(\vec{Q}_k^n).
\end{equation}

\item \label{a:prob_bnd} For any $T > 0$, there exists a finite constant $B(T)$ such that
\[
\lim_{n \to \infty} \PP(\max_{k \le nT}\norm{\vec{Q}_k^n}_\infty > B(T)) = 0,
\]
where $\norm{\cdot}_\infty$ is the $\ell_\infty$ norm of a vector.

\item \label{a:cap} Define 
\[
Q_k^n(\ell; h) = \inprod{\mu_k^n, p_\ell(x, \xi) \sqcap h},
\]
where $\sqcap$ is the projection operation defined in \eref{proj_B}.
For any $b > B(T)$ and $T > 0$, we have
\begin{equation}\label{eq:cap_drift}
\begin{aligned}
&\limsup_{h \to \infty} \, \sup_n \, \max_{k \le nT} \EE \bigg\lvert \mathcal{G}(\xki, \xii, \vec{Q}_k^n) \\
&\qquad\qquad\qquad\qquad- \mathcal{G}(\xki, \xii, \vec{Q}_k^n(h)\sqcap b) \bigg\rvert = 0
\end{aligned}
\end{equation}
and
\begin{equation}\label{eq:cap_diffusion}
\limsup_{h \to \infty} \, \sup_n \, \max_{k \le nT} \EE \abs{\Lambda(\vec{Q}_k^n) - \Lambda(\vec{Q}_k^n(h) \sqcap b) } = 0.
\end{equation}

\item \label{a:gd_bnd} For each $T > 0$, there exists $C(T) < \infty$ such that
\[
\max_{k \le nT} \EE (\Gki)^2 \le C(T)\ \text{and}\ \max_{k \le nT}\EE (\Lki)^2 \le C(T).
\]

\item \label{a:m4} For each $T > 0$, there exists $C(T) < \infty$ such that $\max_{k \le nT} \EE (\Dki)^{4}\le C(T) n^{-2}$, and for any $i \neq j$,
\[
\max_{k \le nT} \EE \abs{\EE_k \left( \Dki- \EE_{k} \Dki \right) \left( \Dkj- \EE_{k} \Dkj \right) }\le C(T) n^{-2}.
\]

\item \label{a:uniqueness} For each $b > 0$ and $T > 0$, the following PDE (in weak form) has a unique solution in $D([0, T], \mathcal{M}(\R^2))$: for all bounded test function $f(x, \xi)  \in \mathcal{C}^{3}( \R^{2})$,
\begin{equation}\label{eq:weak_pde_B}
\begin{aligned}
\inprod{\mu_t, f} &= \inprod{\mu_0, f} + \int_0^t \inprod{\mu_s, \mathcal{G}(x_s, \xi_s, \vec{Q}_s \sqcap b) \tpd{}{x}\!f} \dif s\\
&\qquad+  \frac{1}{2} \int_0^t \inprod{\mu_s, \Lambda(x_{s}, \xi_{s}, \vec{Q}_s \sqcap b) \tpd[2]{}{x}\!f} \dif s,
\end{aligned}
\end{equation}
where $Q_s = [Q_s(1), Q_s(2), \ldots, Q_s(r)]$ with 
\begin{equation}\label{eq:Qs}
\quad Q_s(i) = \inprod{\mu_s, p_i(x, \xi)}.
\end{equation}
\end{enumerate}

Recall the definition of the joint empirical measure $\mu_k^n(x, \xi)$ of $\vx_k$ and $\vxi$ as defined in \eref{mu}. We embed the discrete-time measure-valued process $ \mu_{k}^{n}$ in continuous time by defining 
\[
\mu_{t}^{n} \bydef \mu_{ \left \lfloor nt \right \rfloor }^{n}.
\]
For each $T > 0$, we note that $( \mu_{t}^{n})_{0 \leq t \leq T}$
is a piecewise-constant c\`{a}dl\`{a}g process in $\DMR$.

\begin{theorem}\label{thm:general}
Under \ref{a:exchangeable}--\ref{a:uniqueness}, the sequence of measure-valued processes $ \{( \mu_{t}^{n})_{0 \leq t \leq T} \}_{n}$
converges weakly to a deterministic process $( \mu_{t})_{0 \leq t \leq T}$, which is the unique solution of the following PDE: for all bounded test function $f(x, \xi)  \in \mathcal{C}^{3}( \R^{2})$,
\begin{equation}\label{eq:weak_pde}
\begin{aligned}
\inprod{\mu_t, f} &= \inprod{\mu_0, f} + \int_0^t \inprod{\mu_s, \mathcal{G}(x_s, \xi_s, \vec{Q}_s) \tpd{}{x}\!f} \dif s\\
&\qquad\qquad+  \frac{1}{2} \int_0^t \inprod{\mu_s, \Lambda(\vec{Q}_s) \tpd[2]{}{x}\!f} \dif s,
\end{aligned}
\end{equation}
where $\vec{Q}_s$ is as defined in \eref{Qs}.
\end{theorem}

\begin{remark}
The limiting PDE \eref{weak_pde} is very similar to the one given in \eref{weak_pde_B}. The only difference is that, here in \eref{weak_pde}, we no longer need to enforce a ``cap'' on the order parameters $Q_s$.
\end{remark}

\begin{remark}
$\set{(\mu_t^n)_{0 \le t \le T}}_n$ is a sequence of random elements in $D([0, T], \mathcal{M}(\R^2))$. Precisely speaking, Theorem~\ref{thm:general} states that, as $n \to \infty$, the laws of $\set{(\mu_t^n)_{0 \le t \le T}}_n$ convergences to a $\delta$-measure in $\mathcal{M}\left(D([0, T], \mathcal{M}(\R^2))\right)$ and that $\delta$-measure gives its full weight to $(\mu_t)_{0 \le t \le T}$, the unique solution of the PDE.
\end{remark}

Our proof of Theorem~\ref{thm:general} consists of three main steps, following a classical recipe \cite{Meleard:1987, Sznitman:1991}.  First, we show in \sref{tightness} that the laws of the measure-valued stochastic processes$ \{( \mu_{t}^{n})_{0 \leq t \leq T} \}_{n}$ in $\DMR$ are tight. The tightness implies that any subsequence of $ \{( \mu_{t}^{n})_{0 \leq t \leq T} \}_{n}$ must have a converging sub-subsequence. Second, we prove in \sref{convergence} that any converging subsequence must converge to \emph{a} solution of the PDE (\ref{eq:weak_pde}). Third, we provide in \sref{uniqueness} a set of easy-to-verify conditions that are sufficient to guarantee the uniqueness of the solution of the PDE \eref{weak_pde_B}. This uniqueness property, together with the previous two steps, implies that any subsequence of $\set{(\mu_t^n: 0 < t < T)}_{n}$ has a sub-subsequence converging to a \emph{common} limit. It then follows that the entire sequence must also be converging to that same limit, as otherwise we would have a contradiction.

Before embarking on the three components of the proof, we first establish a consequence of Theorem~\ref{thm:general}. As mentioned in \sref{contributions}, many performance metrics of the algorithms (such as the MSE) are simply linear functionals of the empirical measures [see \eref{loss_function}.] In the following proposition, we show the convergence of such functionals.

\begin{proposition}\label{prop:conv_par}
Let $f(x, \xi)$ be a continuous function such that $\abs{f(x, \xi)} \le C(1 + x^2 + \xi^2)$ for some finite constant $C$. Denote by $\mu^*_t$ the solution of the PDE \eref{weak_pde}. Under the same assumptions of Theorem~\ref{thm:general}, we have
\begin{equation}
\EE \abs{\inprod{\mu_{t}^n, f} - \inprod{\mu^\ast_t, f}} \xrightarrow[]{n \to \infty} 0.
\end{equation}
\end{proposition}
\begin{IEEEproof}
For any $b > 0$, 
\begin{align*}
\EE  \abs{\inprod{\mu_t^n, f} - \inprod{\mu^\ast_t, f}} 
\le & \; \EE \abs{\inprod{\mu_t^n, f} - \inprod{\mu_t^n, f \sqcap b}}\\
&+ \EE \abs{\inprod{\mu_{t}^n, f \sqcap b} - \inprod{\mu^\ast_t, f \sqcap b}} \\
&+ \EE \abs{\inprod{\mu^\ast_t, f \sqcap b} - \inprod{\mu^\ast_t, f}}. 
\end{align*}
Since $\set{\mu_t^n}_n$ converges weakly to $\mu^\ast_t$ and $f \sqcap b$ is a bounded and continuous function, we have
\[
\lim_{n \to \infty} \EE \abs{\inprod{\mu_{t}^n, f \sqcap b} - \inprod{\mu^\ast_t, f \sqcap b}} = 0.
\]
It follows that
\begin{equation}\label{eq:conv_par_1}
\begin{aligned}
\limsup_n \EE & \abs{\inprod{\mu_t^n, f} - \inprod{\mu^\ast_t, f}} \\
\le & \limsup_n \EE \abs{\inprod{\mu_t^n, f} - \inprod{\mu_t^n, f \sqcap b}} 
\\
&+ \EE \abs{\inprod{\mu^\ast_t, f \sqcap b} - \inprod{\mu^\ast_t, f}}.
\end{aligned}
\end{equation}
By exchangeability,
\begin{align}
\EE & \abs{\inprod{\mu_t^n, f} - \inprod{\mu_t^n, f \sqcap b}} \nonumber\\
&\le \EE \abs{f(x_{\lfnt}^1, \xi^1) - f(x_{\lfnt}^1, \xi^1) \sqcap b}\nonumber\\
&\le \frac{\EE f^2(x_{\lfnt}^1, \xi^1)}{b}\label{eq:conv_par_2}\\
&\le \frac{C\EE (1+(x_{\lfnt}^1)^4 + (\xi^1)^4)}{b} \le \frac{C(T)}{b},\label{eq:conv_par_3}
\end{align}
where \eref{conv_par_2} is due to Lemma~\ref{lemma:ind_b} given in Appendix~\ref{appendix:useful_lemma}. Since $\abs{\inprod{\mu_t^\ast, f}} \le C \abs{1 + \inprod{\mu_t^\ast, x^2 + \xi^2}} < \infty$ and $f(x, \xi) \sqcap b \to f(x, \xi)$ pointwisely as $b \to \infty$, we have $\lim_{b \to \infty}  \EE \abs{\inprod{\mu^\ast_t, f \sqcap b} - \inprod{\mu^\ast_t, f}} = 0$. Substituting this and \eref{conv_par_3} into \eref{conv_par_1} and letting $b \to \infty$, we are done.
\end{IEEEproof}

\subsection{Tightness}
\label{sec:tightness}

In what follows, we show that the laws of the measure-valued stochastic processes $\{( \mu_{t}^{n})_{0 \leq t \leq T} \}_{n}$ in $\DMR$ are tight. From Kallenberg \cite[Theorem 16.27, p.\ $\!$324]{Kallenberg:2006}, it is sufficient to show that for any bounded
test function $f \in \mathcal{C}^{3}( \R^{2}, \R)$, the sequence of real-valued
processes $\set{(\inprod{\mu_t^n, f})_{0 \le t \le T}}_n$ 
is tight in $ \mathcal{D}([0,T], \R)$.

\begin{proposition}
For any bounded test function $f \in \mathcal{C}^{3}( \R^{2}, \R)$,
the laws of sequence $\set{(\inprod{\mu_t^n, f})_{0 \le t \le T}}_n$ in ${D}([0,T], \R)$
are tight . Hence, the laws of the sequence $ \{( \mu_{t}^{n})_{0 \leq t \leq T} \}_{n}$ in $ \mathcal{D}([0,T], \mathcal{M}( \R^{2}))$ are tight.
\end{proposition}
\begin{IEEEproof}
According to Billingsley \cite[Theorem 13.2, pp.\,139 - 140]{Billingsley:1999}, this
is equivalent to checking the following two conditions.

1. $ \lim_{b \to \infty} \limsup_{n} \PP( \sup_{t \in[0,T]} \abs{\inprod{ \mu_{t}^{n}, f}} \ge b)=0$,
and

2. for each $ \epsilon$, $ \lim_{ \delta \to0} \limsup_{n} \PP( \omega_{n}^{ \prime}( \delta) \ge \epsilon)=0$. Here, $ \omega_{n}^{ \prime}( \delta)$ is the modulus of continuity
of the function $\inprod{\mu_t^n, f}$ in $D([0,T], \R)$,
defined as 
\[
\omega_{n}^{ \prime}( \delta) \bydef \inf_{ \{t_{i} \}} \max_{i} \sup_{s,t \in[t_{i-1},t_{i})} \abs{ \inprod{\mu_t^n, f} - \inprod{\mu_s^n, f}},
\]
 where $ \left \{ t_{i} \right \} $ is a partition of $ \left[0,T \right]$
such that $ \min_{i} \{t_{i}-t_{i-1} \} \geq \delta$.

Since the test function $f$ is bounded, the first condition is satisfied for any
$b \geq \left \Vert f \right \Vert _{ \infty}$. For the second condition,
we will prove it using the uniform partition $ \{t_{i} \}$. The following
lemma shows that the second condition indeed holds.
\end{IEEEproof}
\begin{lemma}
\label{lem:tight-lem}Let $ \{t_{i} = \frac{iT}{K} \}_{0 \leq i \leq K}$ be a uniform partition of $[0,T]$ into $K$ subintervals. For any $ \epsilon>0$, we have
\[
\begin{aligned}
\lim_{K \to \infty} \limsup_{n \to \infty} \PP \bigg( \max_{1<i \leq K} \sup_{t,s \in[t_{i-1},t_{i}]} \abs{ \inprod{\mu_t^n, f} - \inprod{\mu_s^n, f}} \geq \e \bigg)\\
=0.
\end{aligned}
\]
\end{lemma}
\begin{IEEEproof}
For any $s < t$,
\begin{align*}
\inprod{\mu_t^n, f} - \inprod{\mu_s^n, f}  & =\sum_{k=\ns}^{\nt-1}\left[\inprod{\mu_{k+1}^n, f} - \inprod{\mu_k^n, f} \right]\\
 & =\sum_{k=\ns}^{\nt-1}\frac{1}{n}\sum_{i=1}^{n}\left(f(\xnki,\xii)-f(\xki,\xii)\right).
\end{align*}
Using Taylor series expansion, we get
\[
\begin{aligned}
f(\xnki, \xii)=&f(\xki, \xii)+ \Dki \frac{ \partial}{ \partial x}f(\xki, \xii) \\
 &+ \frac{1}{2} (\Dki)^{2} \frac{ \partial^{2}}{ \partial x^2}f(\xki, \xii)+ r_k^i,
\end{aligned}
\]
where
\begin{equation}
r_k^i = \frac{1}{6} (\Dki)^{3} \frac{ \partial^{3}}{ \partial x^{3}}f(c_k^i, \xii),
\end{equation}
for some $c_k^i \in [\xki, \xnki]$, denotes the higher-order remainder terms.
Then, we decompose $\abs{\inprod{\mu_t^n, f} - \inprod{\mu_s^n, f}}$
into four parts:
\begin{equation}\label{eq:4parts}
\abs{\inprod{\mu_t^n, f} - \inprod{\mu_s^n, f}}= \abs{\sum_{k= \ns}^{ \nt-1}\sum_{ \ell=1}^{4} z_{k, \ell}} \leq \sum_{ \ell=1}^{4} \abs{ \sum_{k= \ns}^{ \nt-1}z_{k, \ell} },
\end{equation}
where 
\begin{align}
z_{k,1}  =& \frac{1}{n^2} \sum_{i=1}^{n} \Gki f^{ \prime}(\xki, \xii)+ \frac{1}{2n^2} \sum_{i=1}^{n} \Lki f^{\prime\prime}(\xki, \xii)\label{eq:z_1}\\
z_{k,2}  =& \frac{1}{n} \sum_{i=1}^{n} \left( \Dki- \EE_{k} \left[ \Dki \right] \right)f^{\prime}(\xki, \xii) \label{eq:z_2} \\
z_{k,3} =& \frac{1}{2n} \sum_{i=1}^{n} \left( (\Dki)^{2}- \EE_{k} \left[ (\Dki)^{2} \right] \right)f^{\prime\prime}(\xki, \xii) \label{eq:z_3} \\
z_{k,4}  =& \frac{1}{n} \sum_{i=1}^{n} \left( \EE_{k} \left[ \Dki \right]- \frac{1}{n} \Gki  \right)f^{\prime}(\xki, \xii)\nonumber\\
& + \frac{1}{2n} \sum_{i=1}^{n} \left( \EE_{k} \left[ (\Dki)^{2} \right]- \frac{1}{n} \Lki  \right)f^{\prime\prime}(\xki, \xii) + r_k^i.\label{eq:z_4}
\end{align}
In \eref{z_1}, we recall that $\Gki$ is the quantity defined in \eref{Gki} and $\Lki$ is defined in \eref{Lki}. We also use $f^{\prime}(\xki, \xii)$ and $f^{\prime\prime}(\xki, \xii)$
as the shorthand notation for $ \frac{ \partial}{ \partial x}f(\xki, \xii)$
and $ \frac{ \partial^{2}}{ \partial x^{2}}f(\xki, \xii)$, respectively.
It follows from the decomposition in \eref{4parts} that
\[
\begin{aligned}
\PP\left(\max_{1<i\leq K}\sup_{t,s\in[t_{i-1},t_{i}]}\abs{\inprod{\mu_t^n, f} - \inprod{\mu_s^n, f}}\geq\e\right) \\
\le L_1 + L_2 + L_3 + L_4,
\end{aligned}
\]
where 
\begin{equation}
L_{ \ell}= \PP \left( \max_{1<i \leq K} \sup_{t,s \in[t_{i-1},t_{i}]} \left| \sum_{k= \ns}^{ \nt-1}z_{k, \ell} \right| \geq \frac{ \epsilon}{4} \right),   \label{eq:L}
\end{equation}
for $ \ell = 1, 2, 3, 4.$
To complete the proof, it is sufficient to show that
\begin{equation}\label{eq:L_lim}
\lim_{K \to \infty} \limsup_{n \to \infty}L_{ \ell}=0
\end{equation}
for $1\le \ell \le 4$. Our strategy is to first show that 
\begin{equation}\label{eq:z_123}
\max_{k \le nT} \EE (z_{k, \ell})^2 \le  C(T) n^{-2}, \quad\text{for } \ell = 1, 2, 3,
\end{equation}
where $C(T)$ is some finite constant that depends on $T$ but not on $n$ nor $k$. Using Lemma~\ref{lem:tight-det} in Appendix~\ref{appendix:useful_lemma}, we can then establish \eref{L_lim} for $\ell = 1, 2, 3$.

To show \eref{z_123} for $ \ell=1$, we note that, for any $k \le nT$, 
\begin{align}
\EE (z_{k,1})^{2}  \leq & \frac{2}{n^{2}} \EE \Big(\frac{1}{n} \sum_{i=1}^{n} \Gki f^{ \prime}(\xki, \xii)\Big)^2 \nonumber\\
& + \frac{1}{2n^2} \EE \Big( \frac{1}{n} \sum_{i=1}^{n} \Lki f^{\prime\prime}(\xki, \xii) \Big)^2\nonumber\\
\leq & \frac{2\norm{f^\prime}_\infty^2}{n^{2}} \EE (\mathcal{G}_k^1)^2 + \frac{\norm{f^{\prime\prime}}_\infty^2}{2n^{2}} \EE (\Gamma_k^1)^2\label{eq:tight_1}\\
\leq & \frac{C(T)}{n^{2}},\label{eq:tight_2}
\end{align}
where in reaching \eref{tight_1} we have used convexity and exchangeability, and \eref{tight_2} follows from assumptions \ref{a:gd_bnd}.

To show \eref{z_123} for $\ell = 2$, we have
\begin{align*}
&\EE (z_{k,2})^{2}\leq    \frac{\norm{f^\prime}_\infty^2}{n^{2}}\sum_{i=1}^{n}\left[\EE(\Dki)^{2}-\EE\left(\EE_{k}(\Dki)^{2}\right)\right]\\
& +\frac{\norm{f^\prime}_\infty^2}{n^2}\sum_{i\neq j}\EE\abs{\EE_k\left(\Dki-\EE_{k}\Dki\right)\left(\Delta_{k,j}-\EE_{k}\Delta_{k,j}\right)}\\
  &\qquad\quad\le  \frac{\norm{f^\prime}_\infty^2}{n}\EE(\D_{k}^1)^{2}
\\
& +\norm{f^\prime}_\infty^2 \EE\abs{\EE_k\left(\Dko-\EE_{k}\Dko\right)\left(\Dkt-\EE_{k}\Dkt\right)}.
\end{align*}
By using H\"{o}lder's inequality $ \EE (\Dki)^{2} \leq \sqrt{ \EE (\Dki)^{4}}$ and assumption~\ref{a:m4}, we can then conclude that $\EE (z_{k,2})^{2} \le C(T) n^{-2}$. The case for $\ell = 3$ can be treated in a similar but more straightforward way:
\begin{align*}
\EE (z_{k,3})^{8}\leq & \frac{\norm{f^\prime}_\infty^2}{2n^{2}}\sum_{i,j=1}^{n}\Big[\EE\left((\Dki)^{2}-\EE_{k}(\Dki)^{2}\right)^2 
\\
&+ \EE\left(\Delta_{k,j}^{2}-\EE_{k}\Delta_{k,j}^{2}\right)^2\Big]\\
\leq  &  \frac{\norm{f^\prime}_\infty^2}{4} \EE (\Dko)^4.
\end{align*}
Using assumption~\ref{a:m4}, we are done. 

The only remaining task is to show \eref{L_lim} for $\ell = 4$. Using assumptions~\ref{a:drift}, \ref{a:diffusion} and exchangeability, we have
\begin{align}
\EE \abs{z_{k, 4}} &\le \frac{\norm{f^\prime}_\infty C(T)}{n^{1+\gamma}} + \frac{\norm{f^{\prime\prime}}_\infty C(T)}{n^{1+\gamma}} + \frac{\norm{f^{\prime\prime\prime}}_\infty}{6}\EE \abs{\Dko}^3 \nonumber\\
&\le \frac{C(T)}{n^{1+\gamma}},\label{eq:tight_3}
\end{align}
for some fixed $\gamma > 0$, where in reaching \eref{tight_3} we have used H\"{o}lder's inequality $\EE \abs{\Dko}^3 \le (\EE (\Dko)^4)^{3/4}$ and assumption~\ref{a:m4}. Markov's inequality gives us
\[
\begin{aligned}
&\PP \left( \max_{1<i \leq K} \sup_{t,s \in[t_{i-1},t_{i}]} \left| \sum_{k= \ns }^{\nt -1}z_{k,4} \right| \geq \frac{\e}{4} \right) \\
&\leq \frac{4}{ \epsilon} \EE \max_{1<i \leq K} \sup_{t,s \in[t_{i-1},t_{i}]} \left| \sum_{k= \ns }^{\nt -1}z_{k,4} \right|\\
&\leq \frac{4}{ \e} \sum_{k=0}^{ \left \lfloor nT \right \rfloor } \EE \left| z_{k, 4} \right| \le \frac{C(T)}{n^\gamma},
\end{aligned}
\]
and thus \eref{L_lim} holds for $\ell = 4$.

\end{IEEEproof}

\subsection{The Limit of Converging Subsequences}
\label{sec:convergence}

In the previous subsection, we have shown the tightness of the laws of the sequence $ \{( \mu_{t}^{n})_{0 \leq t \leq T} \}_{n}$. This property implies the existence of a converging subsequence. 
\begin{proposition}
Let $ \{( \mu_{t}^{n})_{0 \leq t \leq T} \}_{n}$ be a subsequence whose laws $\pi^n$ converges to a limit $\pi^\infty \in \mathcal{M}(\DMR)$. Then $\pi^\infty$ is a Dirac measure concentrated on a deterministic measure-valued process $( \mu_{t})_{0 \leq t \leq T}$, which is the unique solution of the PDE \eref{weak_pde}.
\end{proposition}
\begin{IEEEproof}
We start by showing that $\pi^\infty$ is a Dirac measure concentrated on the unique solution of the regularized PDE given in \eref{weak_pde_B}, provided that the regularization parameter $b$ in \eref{weak_pde_B} satisfies $b \ge B(T)$, where $B(T)$ is the bound given in assumption~\ref{a:prob_bnd}. We then show that all such solutions must also be the unique solution of the original PDE given in \eref{weak_pde}.

For each $t \in (0, T)$ and for each compactly supported function $f(x, \xi) \in C^3(\R^2)$, we define a functional
\begin{equation}\label{eq:def_F}
F: \nu \in \DMR \longmapsto F(\nu), 
\end{equation}
where $F(\nu) = \inprod{\nu_t, f} - \inprod{\nu_0, f} - \int_0^t \inprod{\nu_s, L_{\nu_s} f} \dif s$ and $L_{v_{s}}$ is defined by 
\begin{equation}\label{eq:generator_B}
L_{v_{s}}f= \mathcal{G}(x, \xi, \vec{Q}_s \sqcap b) \frac{ \partial}{ \partial x}f(x, \xi)+ \f 12 \Lambda(\vec{Q}_s \sqcap b) \frac{ \partial^{2}}{ \partial x^{2}}f(x, \xi).
\end{equation}
In the above expression, $\mathcal{G}, \Lambda$ and $\vec{Q}_s$ are defined in assumptions \ref{a:drift}, \ref{a:diffusion} and \eref{Qs}, respectively, and $b > B(T)$ is a fixed constant. To show that $\pi^\infty$ concentrates on the solution of the PDE, it is sufficient to show that
\begin{equation}\label{eq:lim_pi}
\EE_{ \pi^{ \infty}} \abs{F(\nu)}=0.
\end{equation}
To that end, we first establish the following convergence result:
\begin{equation}\label{eq:lim_F}
\lim_{n \to \infty} \EE_{ \pi^{n}} \abs{F(\nu)}=0. 
\end{equation}
Using the decomposition in \eref{4parts}, we have
\begin{align}
&\EE_{\pi^n} \abs{F(\nu)} = \EE 
\left \lvert \sum_{k=0}^{\lfnt} \sum_{\ell = 1}^4 z_{k, \ell} - \f 1n\sum_{k=0}^{\lfnt}\inprod{\muk, L_{\muk}f} \right . \\
&\qquad\qquad \left.- \int_{\frac{\lfnt}{n}}^t \inprod{\mu_s, L_{\mu_s} f} ds \right \rvert
\nonumber
\\
&\le \sum_{k=0}^{\lfnt} \EE \abs{z_{k, 1} - \f 1n \inprod{\mu_k, L_{\mu_k} f}}  + \EE \abs{\sum_{k=0}^{\lfnt} (z_{k, 2} + z_{k, 3})} \nonumber\\
&\qquad\qquad+ \sum_{k=0}^{\lfnt} \EE \abs{z_{k, 4}} + \f 1n \EE \abs{\inprod{\mu_{\lfnt}, L_{\mu_{\lfnt}} f}}.\label{eq:F_1}
\end{align}
The last two terms on the right-hand side of \eref{F_1} can be easily handled. In particular, using the estimate \eref{tight_3} in the proof of Lemma~\ref{lem:tight-lem}, we can bound the third term as
\[
\sum_{k=0}^{\lfnt} \EE \abs{z_{k, 4}} \le \frac{t C(T)}{n^\gamma}.
\]
The fourth term can be shown to be bounded by $C(T)/n$ via exchangeability, assumption~\ref{a:gd_bnd}, and H\"{o}lder's inequality.

Next, we consider the first two terms on the right-hand side of \eref{F_1}. Using the definitions in \eref{z_1} and \eref{generator_B}, we can bound the first term as
\begin{align*}
\sum_{k=0}^{\lfnt}& \EE \abs{z_{k, 1} - \f 1n \inprod{\mu_k, L_{\mu_k} f}} 
\\
\le& t \norm{f^\prime}_\infty \max_{k \le \lfnt} \EE \abs{\mathcal{G}(\xko, \xi^1, \vec{Q}_k) - \mathcal{G}(\xko, \xi^1, \vec{Q}_k \sqcap b)}\\
& + \frac{t \norm{f^{\prime\prime}}_\infty}{2} \max_{k \le \lfnt} \EE \abs{\Lambda(\vec{Q}_k) - \Lambda(\vec{Q}_k \sqcap b)}\\
\le& t \norm{f^\prime}_\infty \PP^{1/2} (\max_{k \le \lfnt} \norm{\vec{Q}_k}_\infty \ge b) \\
& \quad \times \big[\max_{k \le \lfnt} \EE (\mathcal{G}^2(\xko, \xi^1, \vec{Q}_k) 
+ \mathcal{G}^2(\xko, \xi^1, \vec{Q}_k \sqcap b)\big] 
\nonumber\\
&+ \frac{t \norm{f^{\prime\prime}}_\infty}{2} \PP^{1/2} (\max_{k \le \lfnt} \norm{\vec{Q}_k}_\infty \ge b)     \\
&  \quad \times \big[\max_{k \le \lfnt} \EE(\Lambda^2(\vec{Q}_k)+ \Lambda^2(\vec{Q}_k \sqcap b)\big]  ,
\end{align*}
which converges to $0$ as $n \to \infty$ due to assumptions~\ref{a:gd_bnd} and \ref{a:prob_bnd}, for any $b > B(T)$.

To bound the second term on the right-hand side of \eref{F_1}, we note that the sequence $\sum_{k=0}^\ell (z_{k, 2} + z_{k, 3})$ is a martingale with conditionally independent increments. Thus, 
\begin{align*}
\EE \Big(\sum_{k=0}^{\lfnt} (z_{k, 2} + z_{k, 3})\Big)^2 &= \sum_{k=0}^{\lfnt} \EE (z_{k, 2} + z_{k, 3})^2\\
&\le 2 \sum_{k=0}^{\lfnt} \EE(z_{k, 2})^2 + 2 \sum_{k=0}^{\lfnt} \EE(z_{k, 3})^2\\
&\le \frac{C(T)}{n},
\end{align*}
where the last inequality is due to the bound $\max_{k \le nT} \EE (z_{k, \ell})^2 \le  C(T) n^{-2}$ for $\ell = 2, 3$ established in the proof of Lemma~\ref{lem:tight-lem}. Applying H\"{o}lder's inequality, we can thus show that $\EE \abs{\sum_{k=0}^{\lfnt} (z_{k, 2} + z_{k, 3})} \to 0$ as $n \to \infty$.

Now that we have shown that the right-hand side of \eref{F_1} tends to $0$ as $n \to \infty$, we can then establish \eref{lim_F}. To conclude \eref{lim_pi} from \eref{lim_F} via the weak convergence of the sequence $\set{\pi^n}$, the functional $F$ would need to be continuous and bounded. Since the test function $f$ is bounded and compactly supported, $F$ is uniformly bounded. The challenge lies in the generator $L_{\nu_s} f$, defined in \eref{generator_B}. Recall that the order parameter $\vec{Q}_s$ in \eref{generator_B} is a vector whose $i$th component is $Q_s(i) = \inprod{\nu_s, p_i}$, where $p_i(x, \xi)$ is a function that is not necessarily bounded. As a result, $Q_s(i)$ is not necessarily a continuous functional of $\nu$. To overcome this difficulty,
we follow the strategy used in \cite{Jourdain:2015} by defining a modified
operator $L_{ \nu_{t},h}$ as
\begin{equation}\label{eq:generator_rB}
\begin{aligned}
L_{\nu_{s}, h}f= &\mathcal{G}(x, \xi, \vec{Q}_s(h) \sqcap b) \frac{ \partial}{ \partial x}f(x, \xi) \\
&+ \f 12 \Lambda(\vec{Q}_s(h) \sqcap b) \frac{ \partial^{2}}{ \partial x^{2}}f(x, \xi),
\end{aligned}
\end{equation}
where $\vec{Q}_s(h)$ is an $r$-dimensional vector whose $\ell$th element is $Q_s(\ell, h) = \inprod{\nu_s, p_\ell \sqcap h}$. The regularization provided by $h$ ensures that $\vec{Q}_s(h)$ is a continuous functional of $\nu$. Accordingly, we define a modified functional $F_{h}$ similar to \eref{def_F}, with $L_{\nu_s}$ there replaced by $L_{\nu_s, h}$.

Note that $F_{h}$ is uniformly bounded and continuous, and that $F_{h}$ converges pointwisely to $F$ as $h \to \infty$. In addition, $F$ is uniformly bounded. Using \eref{lim_F}, we have 
\begin{align*}
\EE_{\pi^{\infty}} \abs{F(\nu)} & =\lim_{h\to\infty}\EE_{\pi^{\infty}} \abs{F_h(\nu)}\\
 & =\lim_{h\to\infty}\lim_{n\to\infty}\EE_{\pi^{n}} \abs{F_h(\nu)}\\
 & \leq\limsup_{h\to\infty} \, \limsup_{n\to\infty}\EE_{\pi^{n}}\abs{F_h(\nu) - F(\nu)}.
\end{align*}
Note that
\begin{align*}
&\sup_n  \EE_{\pi^{n}}\abs{F_h(\nu) - F(\nu)} \\
\le & \sup_n \f 1n \EE \abs{\sum_{k = 0}^{\lfnt} \inprod{\mu_k^n, L_{\mu_k^n, h} f} - \sum_{k = 0}^{\lfnt}\inprod{\mu_k^n, L_{\mu_k^n} f}}\\
\le & \sup_n   \frac{C}{n} \sum_{k=0}^{\lfnt} \bigg(\EE \abs{\mathcal{G}(\xko, \xi^1, \vec{Q}_k^n\sqcap b) - \mathcal{G}(\xko, \xi^1, \vec{Q}_k^n(h)\sqcap b) } \\
&+ \EE \abs{\Lambda(\xko, \xi^1, \vec{Q}_k^n\sqcap b) - \Lambda(\xko, \xi^1, \vec{Q}_k^n(h)\sqcap b) }\bigg)\\
\le  & CT \sup_n \max_{k \le \lfloor nT \rfloor}  \bigg(\EE \abs{\mathcal{G}(\xko, \xi^1, \vec{Q}_k^n\sqcap b) - \mathcal{G}(\xko, \xi^1, \vec{Q}_k^n(h)\sqcap b) } \\
& + \EE \abs{\Lambda(\xko, \xi^1, \vec{Q}_k^n\sqcap b) - \Lambda(\xko, \xi^1, \vec{Q}_k^n(h)\sqcap b) }\bigg).
\end{align*}
Using assumption~\ref{a:cap}, we are done.
\end{IEEEproof}

\subsection{Uniqueness of the Solution of the Limiting PDE}
\label{sec:uniqueness}

In the statement of Theorem~\ref{thm:general}, the uniqueness of the solution of the PDE \eref{weak_pde_B} is stated as an assumption [see \ref{a:uniqueness}]. In this subsection, we provide a set of easy-to-verify conditions that sufficiently implies \ref{a:uniqueness}. Note that the PDE \eqref{eq:weak_pde_B} is specified by the cap constant $b$ and the following four functions:
\begin{enumerate}[label={(\arabic*)}]
  \item the initial measure $\mu_0$;
  \item the $\R\times\R \to \R^r$ function $\vp(x,\xi)$ that defines the order parameter $\mQ_t=\inprod{\mu_t, \vp(x,\xi)}$;
  \item the drift coefficient $\mathcal{G}(x,\xi,\mQ)$; 
  \item and the diffusion coefficient $\Lambda(\mQ)$.
\end{enumerate}
We assume the following sufficient conditions:
\begin{enumerate}[label={(C.\arabic*)}]
  \item \label{c:init}  $\inprod{\mu_0,\xi^2}\leq L$ and $\inprod{\mu_0,x^2}\leq V$, where $L, V$ are two generic constants.
  \item \label{c:op-lip} For any $x$, $\wt{x}$, and $\xi$, we have $\norm{\vp(x,\xi) - \vp(\wt{x},{\xi}) }_1
    \leq L ( 1+ \abs{x} + \abs{\wt{x}} + \abs{\xi} )\abs{x-\wt{x}}$;
  \item \label{c:op-bd} $\norm{\vp(0,0)}_1\leq L$;
  \item \label{c:drift-lip-x} For any $x$, $\xi$, and $\mQ$, we have $\abs{\Gamma(x,\xi,\mQ) - \Gamma(\wt{x},\xi,\mQ)} \leq L 
    (1+ \norm{\mQ}_1) \abs{(x-\wt{x})}$;
  \item \label{c:drift-lip-op} $ \abs{\Gamma(x,\xi,\mQ) - \Gamma(x,\xi,\wt{\mQ})} \leq L
    (1+\abs{\xi}+\abs{x})\norm{\mQ-\wt{\mQ}}_1$;
  \item \label{c:drift-growth} $\abs{\Gamma(x,\xi,\mQ)}\leq L (1+\abs{\xi} + \abs{x})(\norm{\mQ}_1 + 1) $;
  \item \label{c:diffusion-lip} For any $\mQ$ and $\wt{\mQ}$, we have $\abs{\Lambda^{\tfrac{1}{2}}(\mQ) - \Lambda^{\tfrac{1}{2}}(\wt{\mQ})}\leq L 
    \norm{\mQ - \wt{\mQ}}_1$;
  \item \label{c:diffusion-growth} $\Lambda^{\tfrac{1}{2}}(\mQ)\leq L (1+\norm{\mQ}_1)$.
\end{enumerate}


\begin{theorem} \label{thm:pde-uniq}
  Under \ref{c:init} to \ref{c:diffusion-growth}, and for any finite $T,b>0$, the PDE \eqref{eq:weak_pde_B} has a unique solution $( \mu_t^\ast )_{0\leq t\leq T}$. 
\end{theorem}
\begin{remark}
Our general strategy for proving Theorem~\ref{thm:pde-uniq} is to show that the solution of the PDE is the fixed point of a contraction mapping. To that end, we introduce three mappings. We first define a mapping $\mathcal{F}_{a}:\mathcal D([0,T],\R^{d})\to \mathcal D([0,T],\mathcal{M}(\R^{2}))$
as follows. Let $(\mR_t)_{0\leq t \leq T}\in \mathcal D([0,T],\R^{d})$.
Consider the following standard SDE 
\begin{equation}
\begin{aligned}X_{t} & =X_{0}+\int_{0}^{t}\mathcal{G}(x_{s},\xi_{s},\mR_s \sqcap b)ds+\int_{0}^{t}\Lambda^{\tfrac{1}{2}}(\mR_s \sqcap b)\,dB_{s}\\
\xi_{t} & =\xi_{0},
\end{aligned}
\label{eq:sde-gen}
\end{equation}
where the random variables $(X_{0},\xi_{0})$ are drawn from
the  initial measure $\mu_0$ of the McKean-type PDE \eqref{eq:weak_pde_B}.
Let $(X_{t},\xi_{t})$ be the solution to \eqref{eq:sde-gen}. 
We define $\mathcal{F}_{a}(g)$ as the time-marginal of the law of $(X_{t},\xi_{t})$.

The second mapping $\mathcal{F}_{b}:  \mathcal  D([0,T],\mathcal{M}(\R^{2}))\to \mathcal D([0,T],\R^{d})$
takes a measure-valued process $\mu_{t}(x,\xi)$ and maps it to a
function $(\mS_t)_{0\leq t \leq T}=\mathcal{F}_{b}(\mu)$, defined as 
\[
\mS_t=\inprod{\mu_{t}(x,\xi),\vp(x,\xi)}.
\]
Finally, we consider $\mathcal{F}: \mathcal  D([0,T],\R^{d})\to  \mathcal  D([0,T],\R^{d})$
as a composite of the two mappings, \ie $\mathcal{F}=\mathcal{F}_{b}\circ\mathcal{F}_{a}$.

We note that, if $(\mR_t)_{0\leq t \leq T}$ is a fixed point of $\mathcal{F}$, then
$\mathcal{F}_{a}((\mR_t)_{0\leq t \leq T})$ must be a solution of our PDE \eqref{eq:weak_pde_B}.
Similarly, if $\mu_{t}(x,\xi)$ is a solution of \eqref{eq:weak_pde_B},
then $\mathcal{F}_{b}(\mu)$ must be a fixed point of $\mathcal{F}$.
In Proposition~\ref{prop:contraction} given in Appendix~\ref{appendix:pde}, we show that $\mathcal{F}$ is a contraction when
$T$ is small enough. This then guarantees the existence and 
uniqueness of the solution
of the PDE over the interval $[0,T]$ for a small but finite $T$.
Piecing solutions together, we can then enlarge the interval to the
entire real line.
\end{remark}

\begin{IEEEproof}[Proof of Theorem~\ref{thm:pde-uniq}]
  Proposition~\ref{prop:contraction} in Appendix~\ref{appendix:pde} guarantees the uniqueness of the solution for a small interval $T$.
  If $T$ is large, we need to split the whole  interval $[0, T]$ into $m$ equal-length sub-interval $[T_\ell, T_{\ell+1}]$, $\ell=0,1,\ldots,m-1$ with $T_\ell=\tfrac{\ell}{m}T$. Proposition~\ref{prop:contraction} can then be applied repeatedly, over each sub-interval, as long as the initial variance for each sub-interval stays uniformly bounded, for any $m$.
    For the $\ell$th sub-interval, let its initial variance $\EE X_0^2$ be $V^{(\ell)}$, and we write $V^{(0)}=V$. 
  Next, we are going to show $V^{(\ell)}$ will not diverge as $m\to\infty$, so that we can choose arbitary small length of the sub-intervals.

  From \eqref{eq:op-bd}, we have 
  $$ 
  V^{(\ell+1)}  \leq \left(V^{(\ell)} + {B}(L,db)\tfrac{T}{m} \right) e^{B(L,db) \tfrac{T}{m}},
  $$
  with $\ell=0,1,\ldots,m-1$.
  For any $m>B(L,db)T$, we have
  \begin{align}
  V^{(\ell+1)}  &\leq \left(V^{(\ell)} + {B}(L,db)\tfrac{T}{m} \right)\left(1+2B(L,db) \tfrac{T}{m}\right) \\
   &\leq V^{(\ell)} + {B}(L,db) \left( 3+2 V^{(\ell)} \right)\tfrac{T}{m},
  \end{align}
which implies
  $$
  V^{(\ell)}  \leq  \left( V^{(0)} +\wt{B}(L,db) \right)
   e^{\wt{B}(L,db)T\tfrac{\ell}{m}} \leq C(L,V^{(0)},db,T).
  $$
  This ensures that \ref{c:init} holds for any sub-interval, which concludes the proof.
\end{IEEEproof}

\begin{remark}[Efficient numerical solutions of the PDE]\label{rem:numerical_PDE}
The contraction mapping idea behind the proof of Theorem~\ref{thm:pde-uniq} naturally leads to an efficient numerical scheme to solve the nonlinear PDE \eqref{eq:weak_pde}. Specifically, starting from an arbitrary guess of $\vec{Q}_t^{(0)}$ (for example, a constant function), we solve \eref{weak_pde} by treating it as a standard Fokker-Planck equation with a fixed order parameters $\vec{Q}_t^{(0)}$. Stable and efficient numerical solvers for Fokker-Planck equations are readily available. We can then repeat the process, with a new $\vec{Q}_t^{(k)}$ (for $k \ge 1$) computed from the density solution of the previous iteration by using the formula given in \eref{Qs}. In practice, we find that, by setting the time-interval length to $\Delta T=5$, the algorithm converges in a few iterations in most cases. If the algorithm does not converge, we can reduce the previous length by half and redo the iterations. Extensive numerical simulations show that this method is very efficient in various settings. This simple scheme indicates that solving the nonlinear limiting PDE is no harder than solving a few classical Fokker-Planck equations, making our PDE analysis very tractable.
\end{remark}



\section{The Scaling Limit of Online Regularized Regression: Proofs and Technical Details}
\label{sec:regression_proof}

In this section, we prove Theorem~\ref{thm:lasso}, which provides the scaling limit of the dynamics of online regularized regression algorithm. Since the algorithm is just a special case of the general exchangeable Markov process considered in Theorem~\ref{thm:general}, our tasks here amount to verifying that all the assumptions \ref{a:exchangeable}--\ref{a:uniqueness} of Theorem~\ref{thm:general} hold for the regularized regression algorithm.


\subsection{The Drift and Diffusion Terms}

Recall the definition of the difference term $\Dki$ given in \sref{moment}. Next, we compute the first and second moments of $\Dki$, which correspond to the drift and diffusion terms in the limiting PDE. To that end, we first write
\begin{equation}\label{eq:Dgd}
\Dki = \gki - \frac{\varphi(\xki)}{n} + d_k^i,
\end{equation}
where
\begin{equation}\label{eq:dki}
d_k^i = \frac{\varphi(\xki) - \varphi(\xki + g_k^i)}{n}.
\end{equation} 

\begin{lemma}\label{lemma:gdphi} There exists a finite constant $C$ such that
\begin{align}
\EE (\gki)^2 &\le \frac{C}{n}(1 + \EE\, e_k) \label{eq:g2_bnd}\\
\EE (d_k^i)^2 &\le \frac{C}{n^3}(1 + \EE\, e_k)\label{eq:d_bnd}\\
\EE (\varphi^2(\xki) \vee& \varphi^2(\xki + \gki)) \le C(1 + \EE\, e_k)\label{eq:phi2phig2_bnd}
\end{align}
\end{lemma}
\begin{remark}
Here and throughout the paper, we will repeatedly use in our derivations the following inequality: Let $k$ be a positive integer, and $x_1, x_2, \ldots, x_\ell$ a collection of nonnegative numbers. Then
\begin{equation}\label{eq:convex_ineq}
(x_1 + x_2 + \ldots + x_\ell)^k \le \ell^{k-1} (x_1^k + x_2^k + \ldots + x_\ell^k),
\end{equation}
which follows from the convexity of the function $f(x) = x^k$ on the interval $x \ge 0$.
\end{remark}
\begin{IEEEproof}
Using \eref{g_m2} and exchangeability, we have 
\[
\begin{aligned}
\EE (\gki)^2 &\le \frac{C}{n}(1 + \EE e_k + \EE (\xki - \xi^i)^2) \\
&= \frac{C}{n}(1 + \EE e_k+ \frac{\EE\sum_i (\xki - \xi^i)^2}{n})
\end{aligned}
\]
and thus \eref{g2_bnd}. The second inequality \eref{d_bnd} is an immediate consequence of \eref{g2_bnd} and the Lipschitz continuity of $\varphi(x)$. Next, to verify \eref{phi2phig2_bnd}, we note that the Lipschitz continuity of $\varphi(x)$ implies that $\abs{\varphi(x)} \le C(1+\abs{x})$ for some $C <\infty$. It follows that
\[
\begin{aligned}
\EE\varphi^2(\xki) &\le C \EE (1 + \abs{\xki})^2 \\
&\le C \EE (1 + \abs{\xki - \xi^i} + \abs{\xi^i})^2 \\
&\le 3C (1 + \EE (\xki - \xi^i)^2 + \EE (\xi^i)^2).
\end{aligned}
\]
Using exchangeability and the fact that $\EE (\xi^i)^2 < \infty$, we have
\begin{equation}\label{eq:phi2_bnd}
\EE\varphi^2(\xki) \le C(1 + \EE e_k).
\end{equation}
Finally, we bound $\EE \varphi^2(\xki + \gki)$, we note that
\[
\begin{aligned}
\varphi^2(\xki + \gki) &= (\varphi(\xki) + [\varphi(\xki + \gki) - \varphi(\xki)])^2  \\
&\le C (\varphi^2(\xki) + (\gki)^2).
\end{aligned}
\]
Combining \eref{phi2_bnd} and \eref{g2_bnd}, we are done.
\end{IEEEproof}

\begin{lemma}\label{lemma:m1m2} Let
\begin{equation}\label{eq:drift_diffusion}
\mathcal{G}(x, \xi) \bydef -\tau(x-\xi) - \varphi(x) \quad \text{and} \quad \Lambda(e) = \tau^2(\sigma^2 + e).
\end{equation}
Then
\begin{align}
\EE \abs{\EE_k \Dki - \tfrac{1}{n}\mathcal{G}(\xki, \xi^i)} &\le \frac{C \sqrt{1 + \EE e_k}}{n^{3/2}}\label{eq:m1_gradient}\\
\EE \abs{\EE_k (\Dki)^2 - \tfrac{1}{n}\Lambda(e_k)} &\le  \frac{C (1 + \EE e_k)}{n^{3/2}}\label{eq:m2_diffusion}
\end{align}
\end{lemma}
\begin{IEEEproof}
Using \eref{Dgd} and \eref{g_m1}, we have
\[
\EE \abs{\EE_k \Dki - \tfrac{1}{n}\mathcal{G}(\xki, \xi^i)} = \EE \abs{\EE_k d_k^i} \le \sqrt{\EE (d_k^i)^2},
\]
which, together with \eref{d_bnd}, gives us \eref{m1_gradient}. Using the definition in \eref{Dki}, we can expand the left-hand side of \eref{m2_diffusion} as 
\begin{align}
&\EE \abs{\EE_k (\Dki)^2 - \tfrac{1}{n}\Lambda(e_k)} \nonumber\\
& \qquad= \EE \left \lvert \EE_k (\gki)^2 - \tfrac{1}{n}\Lambda(e_k) - \EE_k (\tfrac{2}{n} \gki \varphi(\xki + \gki) \right.\nonumber\\
& \qquad \qquad \left. - \tfrac{1}{n^2} \varphi^2(\xki + \gki))\right \rvert \nonumber\\
&\qquad\le\frac{C}{n^2} \EE  (\xki - \xi^i)^2 + \frac{2}{n} \EE\abs{\gki \varphi(\xki + \gki)} \nonumber\\
&\qquad \qquad+ \frac{1}{n^2} \EE\varphi^2(\xki + \gki)\label{eq:g_m2_app}\\
&\qquad\le \frac{C}{n^2} \EE e_k + \frac{2}{n} \sqrt{\EE (\gki)^2} \sqrt{\EE \varphi^2(\xki+\gki)} \nonumber\\
&\qquad \qquad + \frac{1}{n^2} \EE\varphi^2(\xki + \gki)\nonumber\\
&\qquad\le \frac{C}{n^2} \EE e_k + \frac{C}{n^{3/2}}(1 + \EE e_k) + \frac{C}{n^2}(1+\EE e_k),\nonumber
\end{align}
where in reaching \eref{g_m2_app} we have used the moment formula in \eref{g_m2}.
\end{IEEEproof}

\begin{remark}
The MSE $e_k$ plays a key role in the above bounds. It is an $\mathcal{O}(1)$ quantity, concentrating around its expectation due to the law of large numbers. Later, we will show that, for any $T > 0$, there exists a finite constant $C(T)$ such that $\EE (e_k)^2 \le C(T)$ for all $n$ and all $k \le nT$.
\end{remark}

Lemma~\ref{lemma:m1m2} essentially derive the leading-order term of $\EE_k \Dki$ and $\EE_k (\Dki)^2$, which are the diffusion coefficient $\mathcal{G}$ and diffusion coefficient $\Lambda$ of the PDE \eqref{eq:weak_pde} respectively.

\subsection{Bounding the MSE and Higher-Order Moments}
The moment bounds in Lemma~\ref{lemma:gdphi} and Lemma~\ref{lemma:m1m2} all involve the MSE $e_k$, defined in \eref{MSE}. In this section, we first bound the $2$nd moment of $e_k$, which then allows us to precisely bound several higher-order moments of the random variable $\Dki$.

\begin{lemma}\label{lemma:gkiDki_m4} There exists a finite constant $C$ such that
\begin{align}
\EE_k (\gki)^4 &\le \frac{C}{n^2}(1 + (e_k)^2)\label{eq:gm4_k}\\
\EE_k (\Dki)^4 &\le \frac{C}{n^2}(1 + (e_k)^2 + (\xki)^4)\label{eq:Dki_k}
\end{align}
\end{lemma}
\begin{IEEEproof}
We first remind the reader that we use $C$ to denote a generic constant, whose exact value can change from line to line in our derivations. From the definition in \eref{gki},
\[
\begin{aligned}
\EE_k (\gki)^4 &= \frac{C}{n^2} \EE_k \Big[(w_k - \tfrac{1}{\sqrt{n}}\va_k^T (\vx_k - \vxi))^4 (a_k^i)^4\Big]\\
	&\le \frac{C}{n^2} \EE_k \Big(\big[(w_k)^4 + \tfrac{1}{n^2} (\va_k^T (\vx_k - \vxi))^4\big] (a_k^i)^4\Big)\\
	&\le \frac{C}{n^2}\Bigg(\EE (w_k)^4 \EE (\a_k^i)^4 \\
	&\qquad+ \frac{1}{n^2} \sqrt{\EE_k (\va_k^T (\vx_k - \vxi))^8 \,\EE_k (a_k^i)^8}\Bigg)\\
	&\le \frac{C}{n^2}\big(1 + \norm{\vx_k - \vxi}^4 / n^2 \big),
\end{aligned}
\]
where the last inequality is based on \eref{clt_bnd} in Lemma~\ref{lemma:clt_bnd}.

To show \eref{Dki_k}, we use the Lipschitz continuity of $\varphi(x)$ which implies that $\abs{\varphi(x)} \le C (1 + \abs{x})$ for some fixed constant $C$. This allows us to bound $\EE_k (\Dki)^4$ as
\begin{align}
\EE_k (\Dki)^4 &\le \EE_k \left(\abs{\gki} + \frac{C(1+\abs{\xki} + \abs{\gki})}{n}\right)^4\nonumber\\
	&\le C \EE_k \big((\gki)^4 + \frac{1}{n^4}(1 + (\xki)^4)\big),\label{eq:Dki_k_1}
\end{align}
where the second inequality is due to \eref{convex_ineq}. Substituting \eref{gm4_k} into \eref{Dki_k_1} then gives us \eref{Dki_k}.
\end{IEEEproof}

\begin{proposition}\label{prop:x4e2}
For any $T > 0$, there exists constants $C_1(T)$ and $C_2(T)$ that depend on $T$ but not on $n$ or $k$ such that
\begin{equation}\label{eq:xm4_bnd}
\max_{1 \le k \le \lfloor nT \rfloor} \EE (\xki)^4 \le C_1(T)
\end{equation}
and
\begin{equation}\label{eq:mse2_bnd}
\max_{1 \le k \le \lfloor nT \rfloor} \EE (e_k)^2 \le C_2(T)
\end{equation}
\end{proposition}
\begin{IEEEproof}
We first note that \eref{mse2_bnd} is a simple consequence of \eref{xm4_bnd}. Indeed,
\[
\begin{aligned}
e_k^2 =& \left(\frac{1}{n}\sum_{1 \le i \le n} (\xki - \xi^i)^2\right)^2 \\
\le &  \frac{1}{n}\sum_{1 \le i \le n} (\xki - \xi^i)^4 \le   \frac{8}{n}\sum_{1 \le i \le n} \Big((\xki)^4 + (\xi^i)^4\Big),
\end{aligned}
\]
where the first inequality is due to convexity and the second is an application of \eref{convex_ineq}. Using exchangeability and the boundedness of $\EE (\xi^i)^4$, we then have
\begin{equation}\label{eq:e2x4}
\EE (e_k)^2 \le 8\Big(\EE (\xki)^4 + \EE (\xi^i)^4\Big).
\end{equation}

Next, we establish \eref{xm4_bnd} by showing the following recursive bound
\begin{equation}\label{eq:x4_recursive}
\EE (\xnki)^4 \le (1+\frac{C_1}{n}) \EE (\xki)^4 + \frac{C_2}{n},
\end{equation}
for some constants $C_1$ and $C_2$. To that end, we use \eref{Dki} to write
\begin{align}
\EE (\xnki)^4 =& \EE (\xki + \Dki)^4\nonumber\\
	=& \EE (\xki)^4 + 4 \EE(\xki)^3 \Dki + 6 \EE (\xki)^2 (\Dki)^2  \nonumber\\
	&+ 4 \EE (\xki) (\Dki)^3 + \EE (\Dki)^4.\label{eq:xm4_exp}
\end{align}
Our strategy is to bound the last four terms of \eref{xm4_exp}. We start with the first term:
\begin{align}
\EE&(\xki)^3 \Dki  \nonumber\\
&= \EE \big[(\xki)^3 \EE_k (\gki)\big] - \tfrac{1}{n}\EE (\xki)^3\varphi(\xki + \gki)\nonumber\\
	&= \tfrac{-\tau}{n}(\EE (\xki)^4 - \EE (\xki)^3 \xi^i) - \tfrac{1}{n}\EE (\xki)^3\varphi(\xki + \gki)\nonumber\\
	&\le \tfrac{-\tau}{n}\EE (\xki)^4 + \tfrac{\tau}{n}[\tfrac{3}{4} \EE (\xki)^4 + \tfrac{1}{4} \EE (\xi^i)^4] + \tfrac{1}{n}(\tfrac{3}{4} \EE(\xki)^4 \nonumber\\
	&\quad + \frac{1}{4}\varphi^4(\xki+\gki))\nonumber\\
	&\le \tfrac{C}{n} \EE (\xki)^4 + \tfrac{C}{n} + \tfrac{C}{n} (1+ \EE (\xki)^4 + \EE (\gki)^4)\nonumber\\
	&\le \tfrac{C}{n} \EE (\xki)^4 + \tfrac{C}{n} + \tfrac{C}{n^3}(1+ \EE (\xki)^4),\label{eq:e2_1term}
\end{align}
where in reaching the last inequality we used \eref{gm4_k} and \eref{e2x4}.
Using Young's inequality, we can bound the remaining three terms as
\begin{align}
&6 \EE (\xki)^2 (\Dki)^2 + 4 \EE (\xki) (\Dki)^3 + \EE (\Dki)^4 \nonumber\\
&\qquad\le \tfrac{3}{n}(\EE (\xki)^4 + n^2 \EE (\Dki)^4) \nonumber \\
& \qquad \quad + \tfrac{1}{n^{3/2}}(\EE (\xki)^4 + 3n^2 \EE (\Dki)^4) + \EE (\Dki)^4\nonumber\\
&\qquad\le \tfrac{C}{n}(1 + \EE (\xki)^4),\label{eq:e2_3term}
\end{align}
where to reach \eref{e2_3term} we have used a combination of \eref{Dki_k} and \eref{e2x4}.
Substituting the bounds \eref{e2_1term} and \eref{e2_3term} into \eref{xm4_exp} then gives us \eref{x4_recursive}. Applying this bound recursively, we get 
\[
\begin{aligned}
\EE (\xki)^4 &\le (1 + \tfrac{C_1}{n})^{k-1} \EE (x_1^i)^4 + \tfrac{C_2}{n}\sum_{\ell = 1}^{k-2}(1+\tfrac{C_1}{n})^\ell\\
	&= (1 + \tfrac{C_1}{n})^{k-1} \EE (x_1^i)^4 + (C_2/C_1) [(1+\frac{C_2}{n})^{k-1} - 1].
\end{aligned}
\]
Since $(1 + \frac{C_1}{n})^{k-1}$ is uniformly bounded for $1 \le k \le \lfloor nT \rfloor$, we have shown \eref{xm4_bnd}.
\end{IEEEproof}

The uniform bounds given in the previous proposition on $\EE (\xki)^4$ and $\EE (e_k)^2$ allow us to derive the following estimates on higher order moments of $\Dki$.

\begin{proposition}\label{prop:m4}
For every $T > 0$, there exists a constant $C(T)$ such that,
\begin{align}
\max_{k \le nT}\EE \, \mathcal{G}^4(\xki, \xi^i) &\le C(T)\label{eq:drift_m4_bnd}\\
\max_{k \le nT}\EE \, \Gamma^2(e_k) &\le C(T),\label{eq:diffusion_m2_bnd}
\end{align}
where $\mathcal{G}(\xki, \xi^i)$ and $\Gamma(e_k)$ are the drift and diffusion terms defined in \eref{drift_diffusion}, respectively. Moreover, if $f(x, \xi)$ is a function such that $\abs{f(x, \xi)} \le L (1 + \abs{x} + \abs{\xi})$ for some constant $L < \infty$, then for all $i \neq j$,
\begin{align}
&\max_{k \le nT}\EE \abs{f(\xki, \xii) f(\xkj, \xij) \EE_k\big[(\Dki - \EE_k \Dki) (\Dkj - \EE_k \Dkj)\big]} \nonumber\\
&\qquad\qquad\le \frac{L^2 C(T)}{n^2}.\label{eq:m2_cross}
\end{align}
\end{proposition}
\begin{IEEEproof}
The first two inequalities can be easily verified by using the bounds in Proposition~\ref{prop:x4e2}. To show \eref{m2_cross}, we first note that we can assume without loss of generality that $L = 1$, in which case $\abs{f(x, \xi)} \le 1 + \abs{x} + \abs{\xi}$. Using the shorthand notation $f^i = \abs{f(\xki, \xii)}, f^j =  \abs{f(\xkj, \xij)}$, we write
\begin{align}
\EE& \abs{f^i f^j \EE_k (\Dki - \EE_k \Dki) (\Dkj - \EE_k \Dkj)} \nonumber \\
&= \EE f^i f^j \abs{\EE_k \Dki \Dkj - \EE_k \Dki \EE_k \Dkj}\nonumber\\
&\le \EE f^i f^j \abs{\EE_k \Dki \Dkj} + \EE \big(f^i\abs{\EE_k \Dki} f^j \abs{\EE_k \Dkj}\big).\label{eq:m2_cross_1}
\end{align}
Next, we bound each of the two terms on the right-hand size of \eref{m2_cross_1}. For the first term, we use \eref{Dgd} and exchangeability to write
\[
\begin{aligned}
\EE &f^i f^j\abs{\EE_k (\Dki \Dkj)} \nonumber \\
&= \EE f^i f^j\abs{\EE_k (\gki - \frac{\varphi(\xki)}{n} + d_k^i)(\gkj - \frac{\varphi(\xkj)}{n} + d_k^j)}\\
&\le \EE f^i f^j\abs{\EE_k (\gki - \frac{\varphi(\xki)}{n})(\gkj - \frac{\varphi(\xkj)}{n})} \\
&\quad + 2\EE f^i f^j\abs{\dki \Dkj} + \EE f^i f^j\abs{\dki \dkj}\\
&\le \EE f^i f^j\abs{\EE_k (\gki \gkj)} + \frac{2}{n} \EE \abs{\EE_k \gki}\abs{f^i f^j\varphi(\xkj)} \\
&\quad + \frac{1}{n^2} \EE (f^i \varphi(\xki))^2 
+\frac{1}{n^2} \EE (f^i f^j)^2 
\\
&\quad + n^2 \EE (\dki \Dkj)^2+ \EE (f^i\dki)^2.
\end{aligned}
\]
Recall the definition of $\dki$ in \eref{dki}. From the Lipschitz continuity of $\varphi(x)$, we have $\abs{\dki} \le C \abs{\gki}/n$, for some constant $C < \infty$. Substituting the explicit formulas in \eref{g_m1} and \eref{g_ij} into the right-hand side of the above inequality gives us 
\[
\begin{aligned}
&\EE f^i f^j\abs{\EE_k (\Dki \Dkj)} \\
&  \le \frac{C \EE f^i f^j\abs{(\xki - \xi^i)(\xkj - \xi^j)}}{n^2}+ \frac{C\EE \abs{(\xki - \xi^i) f^i f^j \varphi(\xkj)}}{n^2}\\
&\quad+ \frac{\EE (f^i\varphi(\xki))^2}{n^2} + \frac{1}{n^2} \EE (f^i f^j)^2 \\
&\quad + C\EE [(\gki)^4 + (\Dki)^4] + \frac{C \EE [(f^i)^4 + (\gki)^4]}{n^2}.\nonumber\\
\end{aligned}
\]
Using the inequality $\abs{x_1 x_2 x_3 x_4} \le \sum_i (x_i)^4 / 4$, the bounds $\abs{f(x, \xi)} \le 1 + \abs{x} + \abs{\xi}$ and $\abs{\varphi(x)} \le C (1 + \abs{x})$, and the previous estimates in Lemma~\ref{lemma:gkiDki_m4}, we conclude that
\begin{equation}\label{eq:m2_cross_2}
\EE f^i f^j\abs{\EE_k (\Dki \Dkj)} \le \frac{ C(T)}{n^2}.
\end{equation}
for all $k \le nT$.

We now bound the second term on the right-hand side of \eref{m2_cross_1}. Using Young's inequality and exchangeability, we have
\[
\begin{aligned}
\EE & f^i\abs{\EE_k \Dki} f^j \abs{\EE_k \Dkj} \nonumber \\
&\le \EE (f^i \EE_k \Dki)^2\nonumber\\
&\le \frac{\EE (f^i)^4}{2n^2} + \frac{n^2}{2} \EE (\EE_k \Dki)^4\nonumber\\
&\le \frac{C(T)}{n^2} + \frac{n^2}{2} \EE (\frac{\mathcal{G}(\xki, \xii)}{n} + \EE_k \dki)^4\\
&\le \frac{C(T)}{n^2} + \frac{C \EE \,\mathcal{G}^4(\xki, \xii)}{n^2} + \frac{\EE (\gki)^4}{n^2}.
\end{aligned}
\]
It then follows from \eref{drift_m4_bnd}, \eref{gm4_k} and the bounds in Proposition~\ref{prop:x4e2} that
\begin{equation}\label{eq:m2_cross_3}
\EE f^i\abs{\EE_k \Dki} f^j \abs{\EE_k \Dkj} \le \frac{ C(T)}{n^2}.
\end{equation}
Substituting \eref{m2_cross_2} and \eref{m2_cross_3} into \eref{m2_cross_1}, we reach the desired bound in \eref{m2_cross}.\end{IEEEproof}

In our proof of the scaling limit, we will also need to use the following result, which shows that the MSE $e_k$ has a finite upper bound with high probability, for any $k \le nT$.

\begin{proposition}\label{prop:ek_pb}
For each $T > 0$, there exists finite constants $B(T)$ and $C(T)$ such that
\begin{equation}\label{eq:ek_pb}
\PP\left(\max_{k \le \lfloor nT \rfloor} e_k > B(T)\right) \le \frac{C(T)}{n}.
\end{equation}
\end{proposition}
\begin{IEEEproof}
See Appendix~\ref{appendix:ek_pb}.
\end{IEEEproof}

\subsection{The Scaling Limit}

As stated earlier, the scaling limit of the online regression algorithm given in Theorem~\ref{thm:lasso} can be obtained as a special case of Theorem~\ref{thm:general}. Indeed, for the online regression algorithm, the function $\mathcal{G}(x, \xi, \vec{Q})$ in \eref{Gki} is $\tau(\xi-x) - \varphi(x)$ and the function $\Lambda(\vec{Q})$ in \eref{Lki} is simply $\tau^2 (\sigma^2 + e)$, with the order parameter $\vec{Q}$ being the scalar MSE $e = \inprod{\mu, (x-\xi)^2}$. The conditions in assumptions \ref{a:drift} and \ref{a:diffusion} are guaranteed by Lemma~\ref{lemma:m1m2} and Proposition~\ref{prop:x4e2}. Assumption \ref{a:prob_bnd} is guaranteed by Proposition~\ref{prop:ek_pb}. To verify assumption~\ref{a:cap}, we note that \eref{cap_drift} is trivially satisfied as the function $\mathcal{G}(x, \xi, \vec{Q})$ in this case does not involve an order parameter. To show \eref{cap_diffusion}, we have
\begin{align*}
\EE& \abs{\inprod{\mu_k^n, (x-\xi)^2} \wedge b - \inprod{\mu_k^n, (x-\xi)^2 \wedge h} \wedge b} \\
&\le \EE \abs{\inprod{\mu_k^n, [(x-\xi)^2 - (x-\xi)^2 \wedge h]}}\\
&= \EE \big[(\xko - \xi^1)^2 - (\xko - \xi^1)^2 \wedge h\big]\\
&\le \EE (\xko - \xi)^4 / h,
\end{align*}
where the last inequality is due to Lemma~\ref{lemma:ind_b}. From Proposition~\ref{prop:x4e2}, the fourth moment $\EE (\xko - \xi)^4$ is bounded and thus we have \eref{cap_diffusion}. Next, assumption~\ref{a:gd_bnd} is guaranteed by Proposition~\ref{prop:m4} via H\"{o}lder's inequality, and assumption~\ref{a:m4} also by Proposition~\ref{prop:m4}. Finally, the existence and uniqueness of the solution of the PDE \eref{weak_pde_B} is guaranteed by Theorem~\ref{thm:pde-uniq}. One can easily check that \ref{c:init} to \ref{c:diffusion-growth} hold for $\mathcal{G}(x,\xi)$ and $\Lambda(e)$, and we omit the straightforward derivations here.


\section{Conclusion}

We presented a rigorous asymptotic analysis of the dynamics of online learning algorithms, and apply the results to regularized linear regression and PCA algorithms. In addition, we provided a  meta-theorem for a general high-dimensional exchangeable Markov chain, of which the previous two algorithms are just special examples. Our analysis studies algorithms through the lens of high-dimensional stochastic processes, and thus it does not explicitly depend on whether the underlying optimization problem is convex or nonconvex. This feature makes our analysis techniques a potentially very useful tool in understanding the effectiveness of using low-complexity iterative algorithms for solving high-dimensional nonconvex estimation problems, a line of research that has recently attracted much attention.

In this work we have only considered the case of estimating a single feature vector. The same technique can be naturally extended to settings involving a finite number feature vectors. Another natural extension is to consider time-varying feature vectors, as in adaptive learning or filtering. Both are left as interesting lines for future investigation.

\appendix

\section*{}

\subsection{Useful Lemmas}
\label{appendix:useful_lemma}

In what follows we state and prove several useful lemmas.

\begin{lemma}
Given an exchangeable Markov chain with states $\vx_k \in \mathcal{S}^{\otimes n}$, the sequence of empirical measures $\set{\mu_k}$ associated with the states forms a measure-valued Markov chain in $\mathcal{M}(\mathcal{S})$.
\end{lemma}
\begin{IEEEproof}
We actually establish a slightly stronger result: Let $\mathcal{D}$ be a Borel set in $\mathcal{M}(\mathcal{S})$. We show that
\begin{equation}\label{eq:sym}
\PP(\mu_{k+1} \in \mathcal{D} \mid \vx_{k}) = \frac{1}{n!} \sum_{\pi \in \Pi_n} \PP(\mu_{k+1} \in \mathcal{D} \mid \pi \circ \vx_k).
\end{equation}
Note that the right-hand side of the above equation is a ``symmetrized'' version of left-hand side, and thus it is a function of the empirical measure associated with $\vx^{k}$. It follows that $\set{\mu_k}_k$ forms a  Markov chain and that
\[
\PP(\mu_{k+1} \in \mathcal{D} \mid \mu_{k}) = \PP(\mu_{k+1} \in \mathcal{D} \mid \vx_{k})
\]
for any $\vx_{k}$ associated with $\mu_{k}$.

To show \eref{sym}, we let $\mathcal{B}_\mathcal{D}$ be the set of points in $\mathcal{S}^{\otimes n}$ whose empirical measures belong to $\mathcal{D}$. Clearly, $\PP(\mu_{k+1} \in \mathcal{S} \mid \vx_{k}) = \PP(\vx_{k+1} \in \mathcal{B}_\mathcal{D} \mid x_{k})$. We note that $\mathcal{B}_\mathcal{D}$ is permutation invariant, \emph{i.e.}, $\pi \circ \mathcal{B}_\mathcal{D} = \mathcal{B}_\mathcal{D}$ for any $\pi \in \Pi_p$. Using this invariance and the exchangeable property of the Markov transition kernel given in \eref{kernel}, we can write
\[
\begin{aligned}
K(\vx_{k}, \mathcal{B}_\mathcal{D}) &= K(\pi \circ \vx_{k}, \pi \circ \mathcal{B}_\mathcal{D})\\
& = K(\pi \circ \vx_{k}, \mathcal{B}_\mathcal{D}) = \frac{1}{n!} \sum_{\pi \in \Pi_n} K(\pi \circ \vx_{k}, \mathcal{B}_\mathcal{D}).
\end{aligned}
\]
\end{IEEEproof}

\begin{lemma}\label{lemma:clt_bnd}
Let $a_1, a_2, \ldots, a_n$ be a sequence of i.i.d. random variables. If $\EE a_i = 0$, $\EE (a_i)^2 = 1$ and $\EE \abs{a_i}^t < \infty$ for some $t > 2$, then
\begin{equation}\label{eq:nrm_ineq}
\EE \left(\frac{\sum_i a_i^2}{n}\right)^{t/2} \le \EE \abs{a_1}^{t}.
\end{equation}
Moreover, for any fixed vector $\vx \in \R^n$,
\begin{equation}\label{eq:clt_bnd}
\EE \bigg\lvert\sum_i x_i a_i\bigg\rvert^t \le C \norm{\vx}^t \EE \abs{a_1}^t,
\end{equation}
where $C$ is a finite constant that can depend on $t$.
\end{lemma}
\begin{IEEEproof}
Inequality \eref{nrm_ineq} is a simple consequence of the convexity of the function $f(x) = x^{t/2}$ on the interval $x \ge 0$. 

Observe that \eref{clt_bnd} holds when $\vx$ is the zero vector. In the following, we assume $\norm{\vx} > 0$ and write $\EE \abs{\sum_i x_i a_i}^t = \norm{\vx}^t \EE \abs{\sum_i \tilde{x}_i a_i}^t$, where $\tilde{x}_i = x_i /\norm{\vx}$. Applying a classical inequality of Rosenthal's \cite{Rosenthal:70} to the sequence of independent random variables $\set{\tilde{x}_i a_i}_i$, we have
\begin{align}
\EE \bigg\lvert \sum_i \tilde{x}_i a_i\bigg\rvert^t &\le C \max\Big\{\sum_i \abs{\tilde{x}_i}^t \EE \abs{a_i}^t, (\sum_i (\tilde{x}_i)^2\EE (a_i)^2)^{t/2}\Big\}\nonumber\\
&\le C \big(\EE \abs{a_1}^t \vee (\EE (a_1)^2)^{t/2}\big),\label{eq:clt_bnd1}
\end{align}
where in reaching \eref{clt_bnd1} we have used the fact that $\sum_i (\tilde{x}_i)^2 = 1$ and thus $\abs{x_i}^t \le (\tilde{x}_i)^2$ for all $t \ge 2$. Finally, applying Jensen's inequality to the right-hand side of \eref{clt_bnd1}, we are done.
\end{IEEEproof}

\begin{lemma}\label{lemma:ind_b}
Let $b > 0 $ be a finite constant and $X$ a random variable with bounded second moment. Then
\[
\EE \abs{X - X \sqcap b} \le \frac{\EE X^2}{b},
\]
where $X \sqcap b$ is the projection operator defined in \eref{proj_B}.
\end{lemma}
\begin{IEEEproof}
Using H\"{o}lder's inequality, we have
\begin{align*}
\EE \abs{X - X \sqcap b} &\le \EE (\abs{X} \charfn_{X^2 \ge b^2})\\
&\le \sqrt{\EE X^2} \sqrt{\PP(X^2 \ge b^2)}.
\end{align*}
Applying Markov's inequality then gives us the desired result.
\end{IEEEproof}

\begin{lemma}
\label{lem:tight-det}Let $(z_{k}^{n})_{k \geq0}$ be a discrete-time stochastic
process parametrized by $n$ and let $ \{t_{i} = \frac{iT}{K} \}_{0 \leq i \leq K}$ be a
uniform partition of the interval $[0,T]$. If $\EE (z_k^n)^2 \le C(T) n^{-2}$ for all $k \le nT$, then for any $ \epsilon>0$, we have 
\[
\lim_{K \to \infty} \limsup_{n \to \infty} \PP \biggl( \max_{1<i \leq K} \sup_{t,s \in[t_{i-1},t_{i}]} \abs{ \sum_{k= \ns}^{ \nt-1}z_{k}^{n} } \geq \e \biggr)=0.
\]
\end{lemma}
\begin{IEEEproof}
It follows from Markov's inequality that
\begin{align}
&\PP \biggl( \max_{1<i \leq K} \sup_{t,s \in[t_{i-1},t_{i}]} \abs{ \sum_{k= \ns}^{ \nt-1}z_{k}^{n} } \geq \e \biggr)\nonumber\\
 &\qquad \leq \frac{1}{ \e} \EE \max_{1<i \leq K} \sup_{t,s \in[t_{i-1},t_{i}]} \abs{ \sum_{k= \ns}^{ \nt-1}z_{k}^{n}} \leq \frac{1}{ \e}M_{n,K,T}, \label{eq:inq-1}
\end{align}
where $M_{n,K,T}= \EE \max_{1<i \leq K} \sum_{k= \left \lfloor nt_{i-1} \right \rfloor }^{ \left \lfloor nt_{i} \right \rfloor -1} \left| z_{k}^{n} \right|.$

For any positive number $B$, we have
\begin{align}
M_{n,K,T} & \leq\EE\max_{1<i\leq K}\sum_{k=\left\lfloor nt_{i-1}\right\rfloor }^{\left\lfloor nt_{i}\right\rfloor -1}[B+(\,\abs{z_{k}^{n}}-\abs{z_k^n} \wedge B)]\nonumber\\
 & \leq \frac{nTB}{K}+\EE\max_{1<i\leq K}\sum_{k=\left\lfloor nt_{i-1}\right\rfloor }^{\left\lfloor nt_{i}\right\rfloor -1}(\,\abs{z_{k}^{n}}-\abs{z_k^n} \wedge B) \nonumber\\
 & \leq \frac{nTB}{K}+\sum_{k=1}^{\lfloor nT \rfloor}\EE(\,\abs{z_{k}^{n}}-\abs{z_k^n} \wedge B).\label{eq:tight_d}
\end{align}
Using Lemma~\ref{lemma:ind_b}, we can bound the expectations on the right-hand side of \eref{tight_d} as
\[
\EE(\,\abs{z_{k}^{n}}-\abs{z_k^n} \wedge B) \le \frac{\EE (z_k^n)^2}{B} \le \frac{C(T)}{Bn^2},
\]
where the second inequality follows from the assumption that $\EE (z_k^n)^2 \le C(T)n^{-2}$. Substituting this bound into \eref{tight_d} gives us
\[
M_{p,K,T} \le \frac{nTB}{K} + \frac{TC(T)}{Bn}.
\]
Choosing $B = \sqrt{K}/n$ and using \eref{inq-1}, we are done.
\end{IEEEproof}

\subsection{Some Lemmas Regarding the Solutions of the PDE}
\label{appendix:pde}

Here we collect some results that are used in establishing the uniqueness of the solution of the PDE given in \eref{weak_pde_B}.
\begin{lemma} \label{lem:sde-gen}
Let $(X_{t},\xi_{t})$ be the strong solution to the SDE \eqref{eq:sde-gen}.
Fix $T>0$. 
For any $0\le s<t\leq T$, 
\begin{equation}
  \EE X_{t}^{2}  \leq  (V + B(L,db)t ) e^{B(L,db)t}\label{eq:sde-bd}
\end{equation}
and
\begin{equation}
\begin{aligned}
&  \EE (X_t - X_s)^2\\
& \leq B(L,db)(t-s) \left[ 1+ (t-s) (V + T) e^{B(L,db)T} \right] \label{eq:sde-cont},
\end{aligned}
\end{equation}
where $B(L,db)$ is some constant dependent on $L$, $b$, and the dimension $d$ of the order parameters.
\end{lemma} 
\begin{IEEEproof}
Using It\^{o}'s formula, we have
\[
\begin{aligned}
X_t^2 = X_0^2 + \int_0^t \left[ 2X_s \mathcal{G} (X_s, \xi_s, \mG_s \sqcap b)
+ \Lambda(\mG_s  \sqcap b) \right]ds
 \\
+ 2\int_0^t X_s \Lambda^\frac{1}{2}(\mQ_s  \sqcap b) dB_s. 
\end{aligned}
\]
It follows that
\[
\begin{aligned}
\EE X_t^2 = \EE X_0^2 + 2\int_0^t \EE X_s \mathcal{G}(X_s, \xi_s, \mG_s  \sqcap b) ds
\\
+ \int_0^t \EE \Lambda(\mG_s  \sqcap b) ds.
\end{aligned}
\]
Using the conditions \ref{c:drift-growth} and \ref{c:diffusion-growth} stated in \sref{uniqueness}, we get
\[
\begin{aligned}
\EE X_t^2 = &\EE X_0^2  + 2L(1+db)\int_0^t \EE \abs{X_t} (1+\abs{\xi} + \abs{X_t})ds
\\
&+ tL^2 (1+db)^2 \\
\leq & \EE X_0^2  + t B(L,db) +B(L,db)\int_0^t \EE X_s^2 ds
\end{aligned}
\]
Next, applying Gr\"{o}nwall's inequality, we get \eqref{eq:sde-bd}.

Finally, combining \ref{c:drift-growth}, \ref{c:diffusion-growth} and \eqref{eq:sde-bd}  with the following inequality
\[
\begin{aligned}
\EE(X_t -X_s)^2 \leq 2(t-s) \EE \int_s^t \mathcal{G}^2 (X_u, \xi_u, \mG_u)du \\
+ 2 \int_s^t \Lambda(\mG_u) dt,
\end{aligned}
\] we can prove \eqref{eq:sde-cont} in a straightforward way.
\end{IEEEproof}

\begin{corollary}
  Let $\mS=\mathcal{F}(\mQ)$. Choose any $T<1/B(L,db)$. We have
  \begin{align}
    \sup_{0\leq t \leq T} \norm{\mS_t}_1 \leq& C(L,V) \label{eq:op-bd} \\
    \sup_{0\leq t \leq T} \norm{\mS_t - \mS_s}_1^2  \leq& C(L,V) (t-s) \nonumber \\
    &\times \left[ 1 + (t-s) T e^{C(L,V) T} \right], \label{eq:op-cont}
  \end{align}
where $C(L,V)$ is some constant dependent on $L$ and $V=\EE X_0^2$.
\end{corollary}
\begin{IEEEproof}
With $T<1/B(L,db)$, \eqref{eq:sde-bd} becomes $EX_t^2 \leq 3 (V+1)$. Using \ref{c:init}, \ref{c:op-lip} and \ref{c:op-bd}, we can prove both \eqref{eq:op-bd} and \eqref{eq:op-cont}.
\end{IEEEproof}
\begin{remark}
After a single map $\mS=\mathcal{F}(\mQ)$, we get a bounded and uniformly continuous function $\mS\in \mathcal{C} ([0,T], \R^d)$ if $T$ is small enough.
Thus, in studying the fixed point of $\mathcal{F}$, we can consider $\mathcal{F}$ as a mapping from $C([0, T], \R)$ to $C([0, T], \R)$. This allows us to use, in what follows, the sup metric in $C([0, T], \R)$, which is easier to work with than the standard metric in $D([0, T], \R)$.
\end{remark}

\begin{proposition}  \label{prop:contraction}
  If $\mR_t$ and $\wt{\mR}_t$ are two functions in $C([0, T], \R)$, then
\begin{equation} \label{eq:contraction-gen}
\begin{aligned}
&\sup_{t \in [0, T]}\norm{(\mathcal{F}\mR)_t - (\mathcal{F}\wt{\mR})_t}_1\\
 &\le A(L,V,db,T) Te^{B(L,V,db)T}\sup_{t \in [0, T]}\norm{\mR_t - \wt{\mR}_t}_1,
\end{aligned}
\end{equation}
where 
\[
A(L,V,db,T)\leq C(L,db) \left[1+\left(V+B(L,db)T\right) e^{B(L,db)T}\right].
\]
Thus, for $T$ sufficiently small, $\mathcal{F}$ is a contraction mapping.
\end{proposition}
\begin{IEEEproof}
We denote by $X_t$ the solution of \eqref{eq:sde-gen} and by $\wt{X_t}$ the solution of the same SDE with $\mR_t$ replaced by $\wt{\mR}_t$. We also couple these two solutions by using the same Brownian motion $B_t$ and the same initial random variables $(X_0,\xi_0)$.
We then have
\[
\norm{(\mathcal{F}b)_t  - \mathcal{\wt{F}}b)_t }_1^2 
\leq \left[ \EE \norm{\vp(X_t , \xi) - \vp(\wt{X_t},\xi)}_1 \right]^2.
\]
Using \ref{c:op-lip}, we have
\[
\begin{aligned}
\norm{(\mathcal{F}b)_t  - (\mathcal{\wt{F}}b)_t }_1^2 
\leq & L^2 \left[ \EE \big(1+\abs{X_t}+\abs{\wt{X_t}}+2\abs{\xi} \big) \right. \\
&\qquad\times \left. \abs{X_t - \wt{X_t}}\right]^2\\
&\leq C(L,db) \EE (X_t -\wt{X}_t)^2,
\end{aligned}
\]
where we use the Cauchy-Schwarz inequality, \eqref{eq:convex_ineq} and \eqref{eq:sde-bd} to reach the last line.

With It\^{o}'s formula, we have
\[
\begin{aligned}
\EE(X_t - \wt{X}_t)^2 =& 
2\int_0^t \EE (X_s - \wt{X_s}) \\
& \quad \times \left( \mathcal{G}(X_s, \xi, \mR_s) -
\mathcal{G}(\wt{X}_s, \xi, {\mR}_s) \right)ds\\
&- 
2\int_0^t \EE (X_s - \wt{X_s}) \\
& \quad \times \left( \mathcal{G}(\wt{X}_s, \xi, \mR_s) -
\mathcal{G}(\wt{X}_s, \xi, \wt{\mR}_s) \right)ds \\
& +
  \int_0^t \left( \Lambda(\mR_t)^\frac{1}{2} -  \Lambda(\wt{\mR}_t)^\frac{1}{2} \right)ds
\end{aligned}
\]

Using Young's inequality, \ref{c:drift-lip-x}, \ref{c:drift-lip-op}, and \ref{c:diffusion-lip}, we get
\[
\begin{aligned}
\EE(X_t - \wt{X}_t)^2 \leq& 3Lt\left(1+L+L^2+L\sup_{0\leq s\leq T} \EE \wt{X}_s^2\right)\\
&\quad \times \sup_{0\leq s \leq T} \norm{\mR - \wt{\mR}}_1\\
&+ \left[ 2 + L^2 (1+db)^2 \right] \int_0^t \EE (X_s - \wt{X}_s)^2 ds,
\end{aligned}\label{eq:cont-1}
\]
which implies \eqref{eq:contraction-gen} by Gr\"{o}nwall's inequality and  \eqref{eq:sde-bd}.

\end{IEEEproof}

\subsection{Proof of Proposition~\ref{prop:ek_pb}}
\label{appendix:ek_pb}

We start by considering the difference term
\[
\begin{aligned}
e_{k+1} - e_k &= \frac{1}{n} \sum_i [(\xki + \Dki - \xi^i)^2 - (\xki  - \xi^i)^2]\\
&= \frac{2}{n}\sum_i \zki \Dki + \frac{1}{n} \sum_i (\Dki)^2,
\end{aligned}
\]
where we use $\zki$ to denote $\xki - \xi^i$ to simplify notation. By introducing 
\begin{equation}\label{eq:ek_pb_bk}
b_k \bydef \frac{2}{n}\sum_i \zki (\Dki - \EE_k \Dki) + \frac{1}{n} \sum_i ((\Dki)^2 - \EE_k (\Dki)^2),
\end{equation}
the above difference term can be written as
\begin{align}
&e_{k+1} - e_k \nonumber \\
&= \frac{2}{n}\sum_i \zki \EE_k\Dki + \frac{1}{n} \sum_i \EE_k (\Dki)^2 + b_k\nonumber\\
&\le \frac{2}{n} \sum_i \zki \EE_k \gki + \frac{2}{n^2} \sum_i \abs{\zki}\abs{\EE_k\varphi(\xki + \gki)} \nonumber\\
&\quad + \frac{2}{n} \sum_i \EE_k (\gki)^2+ \frac{2}{n^3} \sum_i \EE_k \varphi^2(\xki + \gki) + b_k\nonumber\\
&\le \frac{-2\tau}{n} e_k + \frac{2}{n} \sqrt{\sum_i (\zki)^2/n} \sqrt{\sum_i \EE_k \varphi^2(\xki+\gki)/n}\nonumber\\
&\quad+\frac{2\tau^2(\sigma^2 + e_k)}{n} + \frac{C}{n^2} e_k + \frac{2}{n^3} \sum_i \EE_k \varphi^2(\xki + \gki) + b_k,\label{eq:ek_pb1}
\end{align}
where in reaching the last inequality we have used the explicit calculations given in \eref{g_m1} and \eref{g_m2}. The Lipschitz continuity of $\varphi(x)$ implies that $\abs{\varphi(x)} \le C(1 + \abs{x})$ for some finite $C$ and thus $\varphi^2(x) \le 2C(1 + x^2)$. It follows that
\[
\begin{aligned}
\frac{1}{n}&\sum_i \EE_k \varphi^2(\xki + \gki) \\
&\le \frac{C}{n} \sum_i (1 +  (\xki)^2 + \EE_k (\gki)^2)\\
&\le \frac{C}{n}\sum_i (1 + 2  (\xki - \xi^i)^2 + 2 (\xi^i)^2 + \EE_k (\gki)^2)\\
&\le C (1 + e_k).
\end{aligned}
\]
Substituting this bound into \eref{ek_pb1} and using the simple inequality $\sqrt{x(1+x)} \le x + 1/2$, we get
\begin{align}
e_{k+1} &\le e_k - \frac{-2\tau}{n}e_k + \frac{C}{n}\sqrt{e_k(1+e_k)} \nonumber \\
&\quad + \frac{C(1+e_k)}{n} + \frac{C}{n^2}(1+e_k) + b_k \nonumber\\
&\le (1 + \frac{C}{n})e_k + \frac{C}{n} + b_k.\label{eq:ek_pb2}
\end{align}
Iterating \eref{ek_pb2} leads to 
\begin{equation}\label{eq:ek_pb3}
e_k \le \gamma^k (e_0 + 1) + s_k,
\end{equation}
for each $k \ge 1$, where $\gamma = (1+ \frac{C}{n})$ and
\[
s_k \bydef \sum_{\ell=0}^{k-1} \gamma^{k-\ell-1} b_{\ell}.
\]
For any fixed $T > 0$, there exists a constant $C(T) < \infty$ such that $\gamma^k < C(T)$ for all $k \le nT$. Choosing $B(T) = 2C(T) (e_0 + 1)$. It then follows from \eref{ek_pb3} that
\begin{equation}\label{eq:ek_pb_p1}
\PP(\max_{0 \le k \le \lfloor nT \rfloor} e_k > B(T)) \le \PP(\max_{0 \le k \le \lfloor nT \rfloor} s_k > \frac{B(T)}{2}).
\end{equation}
By the definition of $b_k$ in \eref{ek_pb_bk}, we can see that $s_k$ is a martingale. Doob's maximum inequality gives us
\begin{align}
\PP(\max_{k \le \lfloor nT \rfloor} s_k > B(T)/2) & \le \frac{4\EE (s_{\lfloor nT \rfloor})^2}{B^2(T)} \nonumber \\
&= \frac{4\sum_{k = 1}^{\lfloor nT \rfloor} \gamma^{2(\lfloor nT \rfloor - k - 1)}\EE b_{k}^2}{B^2(T)}\nonumber\\
&\le C(T)\sum_{k = 1}^{\lfloor nT \rfloor} \EE \, b_{k}^2. \label{eq:ek_pb_p2}
\end{align}
Using the definition of $b_k$ in \eref{ek_pb_bk} and exchangeability, we have
\begin{align}
\EE b_k^2 &\le \frac{C}{n^2} \sum_i \EE \big[(\zki)^2 \EE_k (\Dki - \EE_k \Dki)^2\big] \nonumber\\
&\qquad+ \frac{C}{n^2} \sum_{i \neq j} \EE \big[\zki \zkj \EE_k (\Dki - \EE_k \Dki) (\Dkj - \EE_k \Dkj)\big]\nonumber\\
&\qquad + \frac{C}{n^2} \EE \big( \sum_i [(\Dki)^2 - \EE_k (\Dki)^2]\big)^2\nonumber\\
&\le \frac{C}{n} \EE \big[(\zki)^2 \EE_k (\Dki)^2\big] + C\EE (\Dki)^4\nonumber \\
&\qquad + C \EE \abs{\zki \zkj \EE_k (\Dki - \EE_k \Dki) (\Dkj - \EE_k \Dkj)} \nonumber \\
&\le \frac{C}{n} \EE \big[(\zki)^2 \EE_k (\Dki)^2\big] + \frac{C(T)}{n^2},\label{eq:ek_pb4}
\end{align}
where in reaching the last inequality we have used the bound \eref{m2_cross} by choosing $f(\xki, \xii) = \zki = \xki - \xii$, the bound \eref{Dki_k} and Proposition~\ref{prop:x4e2}. We just need to bound the first term on the right-hand side of \eref{ek_pb4}.
\begin{align}
\frac{1}{n}& \EE  \big[(\zki)^2 \EE_k (\Dki)^2\big] \nonumber \\
&\le \frac{\EE (\zki)^4}{2n^2} + \frac{1}{2}\EE \big[\EE_k (\Dki)^2\big]^2\nonumber\\
&\le \frac{C(T)}{n^2} + C \EE\bigg[\EE_k \bigg(\gki - \frac{\varphi(\xki + \gki)}{n}\bigg)^2\bigg]^2\nonumber\\
&\le \frac{C(T)}{n^2} + C \EE \bigg[\EE_k (\gki)^2 + \frac{1 + (\xki)^2}{n^2}\bigg]^2\nonumber\\
&\le \frac{C(T)}{n^2} + \frac{C}{n^2} \EE \bigg(1 + e_k + \big[(\zki)^2 + 1 + (\xki)^2\big]/n\bigg)^2\label{eq:ek_pb5}\\
&\le \frac{C(T)}{n^2}.\label{eq:ek_pb6}
\end{align}
In reaching \eref{ek_pb5} we have used the explicit moment calculations in \eref{g_m2}. And \eref{ek_pb6} is due to Proposition~\ref{prop:x4e2}. Substituting \eref{ek_pb6} into \eref{ek_pb4}, we can conclude that $\EE b_k^2 \le C(T)/n^2$. It follows from \eref{ek_pb_p1} and \eref{ek_pb_p2} that
\[
\PP\Big(\max_{0 \le k \le \lfloor nT \rfloor} e_k > B(T)\Big) \le \frac{C(T)}{n}.
\]

\bibliographystyle{IEEEtran}
\bibliography{refs,library}

\end{document}